\documentclass[10pt,twocolumn,letterpaper]{article}

\usepackage{cvpr}              % To produce the CAMERA-READY version
\usepackage{bm}
\usepackage{multirow}
\usepackage{capt-of}
\usepackage{flushend, cuted}
\usepackage[table]{xcolor}

\makeatletter
\newcommand{\printfsymbol}[1]{%
\textsuperscript{\@fnsymbol{#1}}%
}
\makeatother

\definecolor{cvprblue}{rgb}{0.21,0.49,0.74}
\usepackage[pagebackref,breaklinks,colorlinks,citecolor=cvprblue]{hyperref}

\def\paperID{4630} % *** Enter the Paper ID here
\def\confName{CVPR}
\def\confYear{2024}

\title{CoSeR: Bridging Image and Language for Cognitive Super-Resolution}
\author{
Haoze Sun\textsuperscript{1} \quad Wenbo Li\textsuperscript{2}\printfsymbol{1} \quad Jianzhuang Liu\textsuperscript{2} \quad Haoyu Chen\textsuperscript{3} \quad \\
Renjing Pei\textsuperscript{2} \quad Xueyi Zou\textsuperscript{2} \quad Youliang Yan\textsuperscript{2} \quad Yujiu Yang\textsuperscript{1}\thanks{Corresponding author} \\
${^1}$Tsinghua University \quad ${^2}$Huawei Noah’s Ark Lab \quad ${^3}$HKUST (GZ) \\
{\tt\small shz22@mails.tsinghua.edu.cn} \quad
{\tt\small liwenbo50@huawei.com}
}

\begin{document}
\maketitle

\begin{strip}\centering
\vspace{-40pt}
\includegraphics[width=\textwidth]{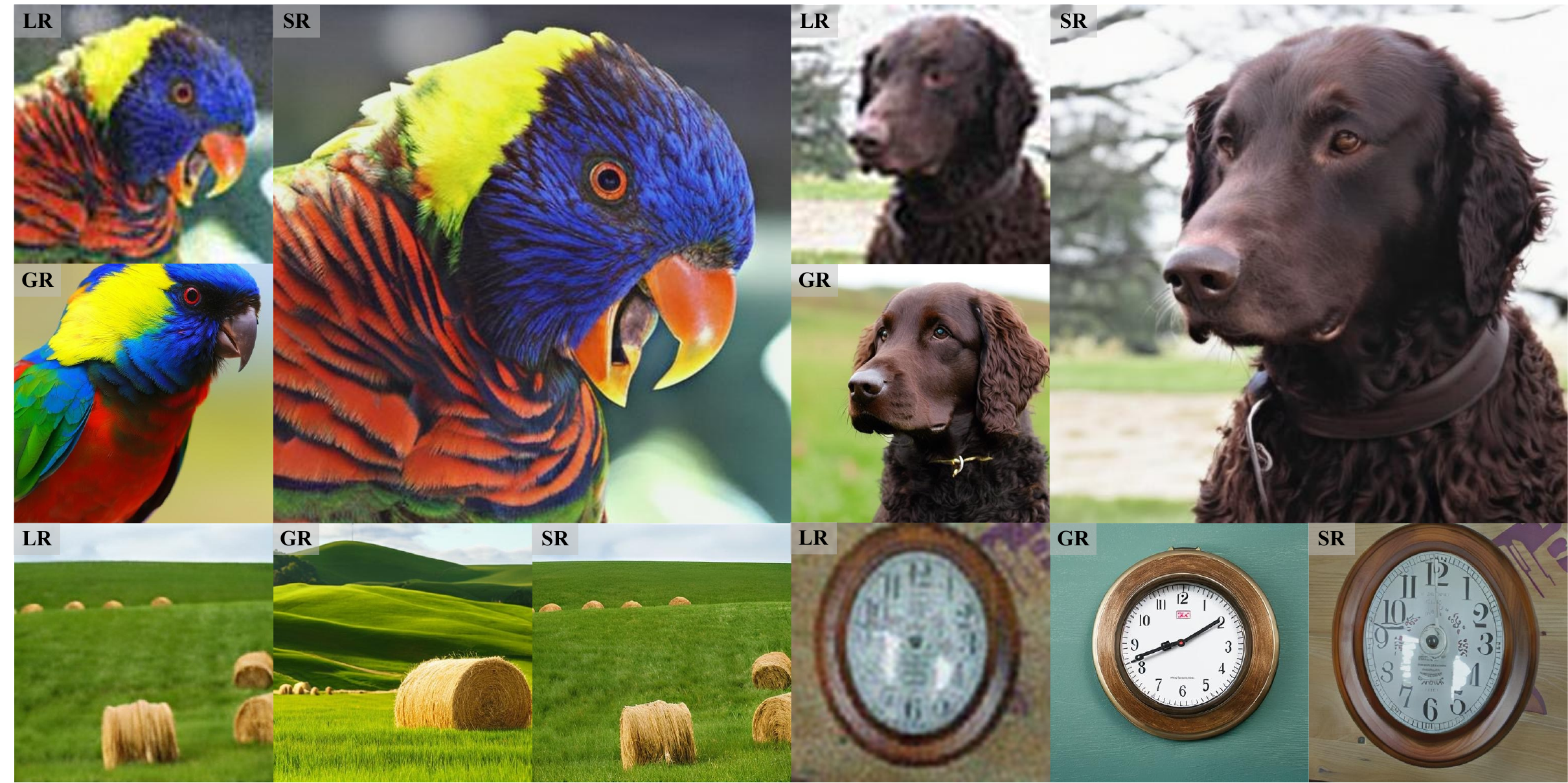}
\captionof{figure}{Visualization displaying 4$\times$ super-resolution results generated by our Cognitive Super-Resolution (CoSeR) model. CoSeR adeptly extracts cognitive information from a low-resolution (LR) image and utilizes it to generate a high-quality reference image. This reference image, aligning closely with the LR image in terms of semantics and textures, significantly benefits the super-resolution process. For conciseness, we denote the input, generated reference, and restoration result as LR, GR, and SR, respectively. Best viewed zoomed in.}\label{fig:teaser}
\end{strip}

% \twocolumn[{%
%    \renewcommand\twocolumn[1][]{#1}%
%    \maketitle
%    \vspace{-5mm}
%   %teas
%    \begin{center}
%     \centering
%     \includegraphics[width=\textwidth]{teaser_v3.pdf}
    
%     % \vspace{-6mm}
%     % \fbox{\rule{0pt}{3cm} \rule{1\linewidth}{0pt}}
%     % \captionof{figure}{Our method can extract cognitive embedding from a very low-resolution input image, which implies an understanding of the scene content. The cognitive embedding can be employed to produce a high-resolution reference image that share semantic information with the input. Finally, we utilize the cognitive embedding to activate the diffusion prior, which, in conjunction with the generated reference image, guides the image restoration process. We show some examples of generated reference images and restoration results.} 
%     \captionof{figure}{Visualization displaying 4$\times$ super-resolution results generated by our Cognitive Super-Resolution (CoSeR) model. CoSeR adeptly extracts cognitive information from a low-resolution (LR) image and utilizes it to generate a high-quality reference image. This reference image, aligning closely with the LR image in terms of semantics and textures, significantly benefits the super-resolution process. For conciseness, we denote the input, generated reference, and restoration result as LR, GR, and SR, respectively. Best viewed zoomed in.}\label{fig:teaser}
%     \vspace{1mm}
%    \end{center}%
% }]

\begin{abstract}
Existing super-resolution (SR) models primarily focus on restoring local texture details, often neglecting the global semantic information within the scene. This oversight can lead to the omission of crucial semantic details or the introduction of inaccurate textures during the recovery process. In our work, we introduce the Cognitive Super-Resolution (CoSeR) framework, empowering SR models with the capacity to comprehend low-resolution images. We achieve this by marrying image appearance and language understanding to generate a cognitive embedding, which not only activates prior information from large text-to-image diffusion models but also facilitates the generation of high-quality reference images to optimize the SR process. To further improve image fidelity, we propose a novel condition injection scheme called ``All-in-Attention'', consolidating all conditional information into a single module. Consequently, our method successfully restores semantically correct and photorealistic details, demonstrating state-of-the-art performance across multiple benchmarks. 
Code: \href{https://github.com/VINHYU/CoSeR}{https://github.com/VINHYU/CoSeR}

% We achieve this by marrying specific image features and generalized language understanding to generate a cognitive embedding for comprehensive scene cognition. This embedding not only extracts prior information from large text-to-image diffusion models but also facilitates the generation of high-quality reference images to optimize the SR process.
\end{abstract}
    
\section{Introduction}

Real-world image super-resolution (SR) is a fundamental task in the realm of image processing, aimed at enhancing low-resolution (LR) images to yield the high-resolution (HR) counterparts~\cite{blind_survey}. Its versatile applicability spans critical domains, including mobile phone photography~\cite{chen2019camera}, autonomous driving~\cite{li2023azimuth}, and robotics~\cite{wang2021real}, while also influencing various computer vision tasks, notably object detection~\cite{haris2021task}, segmentation~\cite{wang2020dual} and recognition~\cite{gunturk2003eigenface,chen2020identity}.

Despite significant advancements in this field in recent years, the processing of complex real-world scenarios continues to pose enduring challenges. Utilizing image priors is a common strategy for tackling real-world SR problems. These priors may be either introduced explicitly in the form of reference images~\cite{ref-transformer,jiang2021robust,zhang2019image,zheng2018crossnet,zhou2020cross,ttsr,lu2021masa, cao2022reference, xia2022coarse}, or implicitly leveraged through pre-trained generative models~\cite{glean, gu2020image, menon2020pulse, pan2021exploiting, wang2021towards, yang2021gan, gdp, ddrm, ddnm, difface, stablesr, pixel-aware, diffbir}. Especially, the recently emerged text-to-image diffusion models~\cite{stablediffusion, sdxl, dall2, imagen} exhibit a remarkable capability to generate high-quality images based on user-provided prompts. These models not only possess strong image priors but also allow precise responses to human instructions in the form of language. This opens up the potential to bridge low-level image processing and high-level abstract cognition.

Consider the process by which human experts restore low-quality images~\cite{nazeri2019edgeconnect, sftgan}: They start by establishing a comprehensive understanding of the image, encompassing scene identification and primary subject recognition. Subsequently, their focus shifts to a meticulous examination and restoration of finer image details. In contrast, conventional image super-resolution techniques~\cite{srgan, rcan, esrgan, swinir, stablesr, diffbir}, adhere to a bottom-up approach, primarily concentrating on local content and direct pixel-level processing. Consequently, these methodologies exhibit inherent limitations in grasping the holistic image context, often failing to restore severely degraded yet semantically vital details. Moreover, given the ill-posed nature of LR images, there is a possibility for introducing semantically erroneous textures~\cite{sftgan}. 

To surmount these challenges, there arises a compelling rationale for imbuing the SR model with ``cognitive" capabilities. In this pursuit, we introduce a pioneering SR methodology known as Cognitive Super-Resolution (CoSeR). Our approach aligns with the top-down cognitive process employed by humans in image perception. It commences with the generation of cognitive embeddings, a representation that encapsulates the overarching comprehension of the LR image, containing both scene semantics and image appearance. This cognitive embedding allows us to precisely leverage the implicit prior knowledge embedded in pre-trained text-to-image generation models, resulting in an enhanced capacity to restore image details in a manner akin to human expertise. Previous work~\cite{sftgan} uses segmentation maps to offer semantics. However, acquiring ideal segmentation maps for real-world LR images remains difficult, even with advanced models like~\cite{sam}. Moreover, semantic segmentation is constrained by predefined categories, limiting its applicability to open-world scenes.

% We opt to model the cognitive process using natural language for two key reasons. Firstly, recent breakthroughs in large language models (LLMs)~\cite{llama2, alpaca, gpt4, palm2} underscore their capacity to effectively approximate human cognitive processes. Our study demonstrates that cognitive embeddings can comprehensively describe LR images (See Figure~\ref{fig:teaser}). Secondly, adopting natural language allows us to precisely leverage the implicit prior knowledge embedded in pre-trained text-to-image generation models. In contrast, the line of using segmentation maps~\cite{sftgan} to offer semantics may encounter difficulties when applied to broader real-world scenarios. On the one hand, acquiring ideal segmentation maps for real-world LR images remains difficult, even with advanced models like~\cite{sam}. Moreover, semantic segmentation is constrained by predefined categories, limiting its applicability to open-world scenes.

Apart from implicitly leveraging diffusion priors, we also advocate for the explicit utilization of image priors. We introduce a novel approach where we employ cognitive embeddings derived from LR inputs to generate reference images through diffusion models. These reference images are subsequently utilized to guide the restoration process. 
% As shown in Figure~\ref{fig:teaser}, our model is capable of generating high-quality reference images that closely resemble the semantic content of the originals. 
As shown in Figure~\ref{fig:teaser}, our cognitive embedding contains language understanding while preserving the color and texture information of the image, thus producing high-quality reference images that are not only semantically aligned but also similar in appearance. 
This explicit approach brings substantial improvements in capturing high-definition textures compared to relying solely on implicit diffusion knowledge. 
% Moreover, unlike earlier reference image-based SR techniques, our method eliminates the laborious reference image search process, enhancing its applicability to broader real-world scenarios.

We have established both implicit and explicit cognitive priors for LR inputs. Then incorporating these priors effectively into our model is pivotal. Unlike the typical conditional generation methods~\cite{controlnet, t2i-adapter, composer, uni-controlnet}, super-resolution demands a heightened level of fidelity between outputs and low-quality inputs. In order to concurrently ensure texture realism and fidelity, we introduce an ``All-in-Attention" design, which integrates multiple information sources via an attention mechanism, including cognitive embeddings, reference images, and LR inputs. This approach allows our model to flexibly use different conditional components, yielding improved results. Our experiments show that our model excels in preserving fidelity compared to previous methods while generating more intricate textures.

The contributions of this paper can be summarized as:
\begin{itemize}
\item We introduce CoSeR, a novel framework for high-detail image super-resolution. CoSeR autonomously extracts cognitive embeddings from LR images, harnessing implicit diffusion priors to enhance the LR input. 
% Additionally, our method accommodates user-provided prompts to customize the scenario cognition.
\item We incorporate diffusion priors explicitly by creating semantically coherent reference images, which act as guidance to improve the quality of the restored image. 
% Furthermore, our approach allows for the inclusion of user-defined reference images.
\item To enhance image fidelity, we introduce a novel ``All-in-Attention'' architecture to integrate conditional information into the SR model. Our method achieves state-of-the-art performance across multiple benchmarks.

\end{itemize}

\section{Related work}
\subsection{Real-World Image Super-Resolution}

Real-world image SR has primarily revolved around two avenues: data utilization and image prior incorporation.

The first category involves the creation of diverse and realistic pairwise data by adapting the physical collection means~\cite{city100, realsr, drealsr} or improving the generation pipeline~\cite{degradation-gan, real-esrgan, bsrgan, yang2023synthesizing}. Also, several works~\cite{cincgan, wei2021unsupervised, maeda2020unpaired} combine both paired and unpaired data with weak supervision to enhance performance in real-world scenarios. 
% However, transitioning between generated and real data domains remains a challenge for existing methods~\cite{blind_survey}.

The second line focuses on the use of image priors. While the ``learning-from-scratch'' approaches~\cite{real-esrgan, bsrgan, sr3} demand substantial data and computational resources, using pre-trained generative models with rich texture priors has become a practical and economical practice. Several studies~\cite{glean, gu2020image, menon2020pulse, li2022best, pan2021exploiting, wang2021towards, yang2021gan} have leveraged pre-trained Generative Adversarial Networks (GANs) to improve the super-resolution process. Nonetheless, these methods occasionally suffer from the generation of unrealistic textures, owing to the inherent limitations of GANs~\cite{liang2022details, DeSRA}. Consequently, there is a growing interest in utilizing more advanced pre-trained generative models, such as the denoising diffusion models~\cite{ddpm, ddim}, in recent research.

\subsection{Diffusion-Based Super-Resolution}
%Some recent real-world image SR methods harness pre-trained diffusion models~\cite{guideddiffusion} but lack text control~\cite{ddrm, ddnm, gdp, difface}. These techniques can leverage the rich texture information in pre-trained models but are usually limited to handling non-blind degradation~\cite{ddnm} or specific scenarios, such as facial images~\cite{difface}. An alternative strategy proposed by Fei et al.~\cite{gdp} concurrently estimates the degradation model to address blind degradation. However, this method is primarily suitable for addressing linear degradation, posing limitations in the face of higher levels of real-world degradation.

Recent approaches~\cite{ddrm, ddnm, difface} utilize implicit knowledge from pre-trained diffusion models~\cite{guideddiffusion}, yet they typically focus on non-blind degradation~\cite{ddnm} or specific domains like facial images~\cite{difface}. In an alternative strategy proposed by Fei \etal~\cite{gdp}, the simultaneous estimation of the degradation model is applied to address blind degradation. However, this method relies on test-time optimization and primarily explores SR under linear degradation, thereby exhibiting limitations in handling real-world complexities.

% Another approach to addressing super-resolution (SR) challenges involves harnessing the recent advancements in large text-to-image diffusion models~\cite{stablediffusion, sdxl, dall2, imagen}. These models, trained on billions of high-definition images, offer improved capabilities for processing diverse content within generic real-world scenarios. StableSR~\cite{stablesr} is the pioneering work that attempts to apply text-to-image priors to real-world SR. It presents a framework for amalgamating prior information from diffusion models and low-resolution (LR) image features within the VQ-GAN~\cite{vqgan} decoder, consequently enhancing fidelity. DiffBIR~\cite{diffbir} integrates the traditional pixel regression-based image recovery model and the text-to-image diffusion model, reducing the deleterious impact of LR degradation on the generation model. PASD~\cite{pixel-aware} prioritizes enhancing the pixel-level quality of realistic SR by introducing pixel-level conditional information to the diffusion model. While these methods have demonstrated substantial improvements in visual quality compared to earlier approaches, they have yet to fully exploit the potential of large text-to-image generation models. Specifically, they have not effectively leveraged the model's capability to extract high-definition texture information corresponding to textual input.

Other approaches~\cite{stablesr, diffbir, pixel-aware} leverage recent advancements in large-scale text-to-image diffusion models~\cite{stablediffusion, sdxl, dall2, imagen}. These models, trained on extensive datasets of high-definition images, provide enhanced capabilities for processing diverse content. StableSR~\cite{stablesr} stands as a pioneering work, which harnesses prior information from diffusion models, resulting in improved fidelity. DiffBIR~\cite{diffbir} combines a traditional pixel regression-based image recovery model with the text-to-image diffusion model, mitigating the adverse effects of LR degradation on the generation process. 
% In contrast, PASD~\cite{pixel-aware} emphasizes enhancing the pixel-level quality by introducing pixel-level conditions to the diffusion model. 
Despite notable advancements in visual quality, these methods have yet to fully harness the potential of large text-to-image generation models, mainly due to the limited image content comprehension.

\begin{figure*}[!htp]
    \centering
    \includegraphics[width=\textwidth]{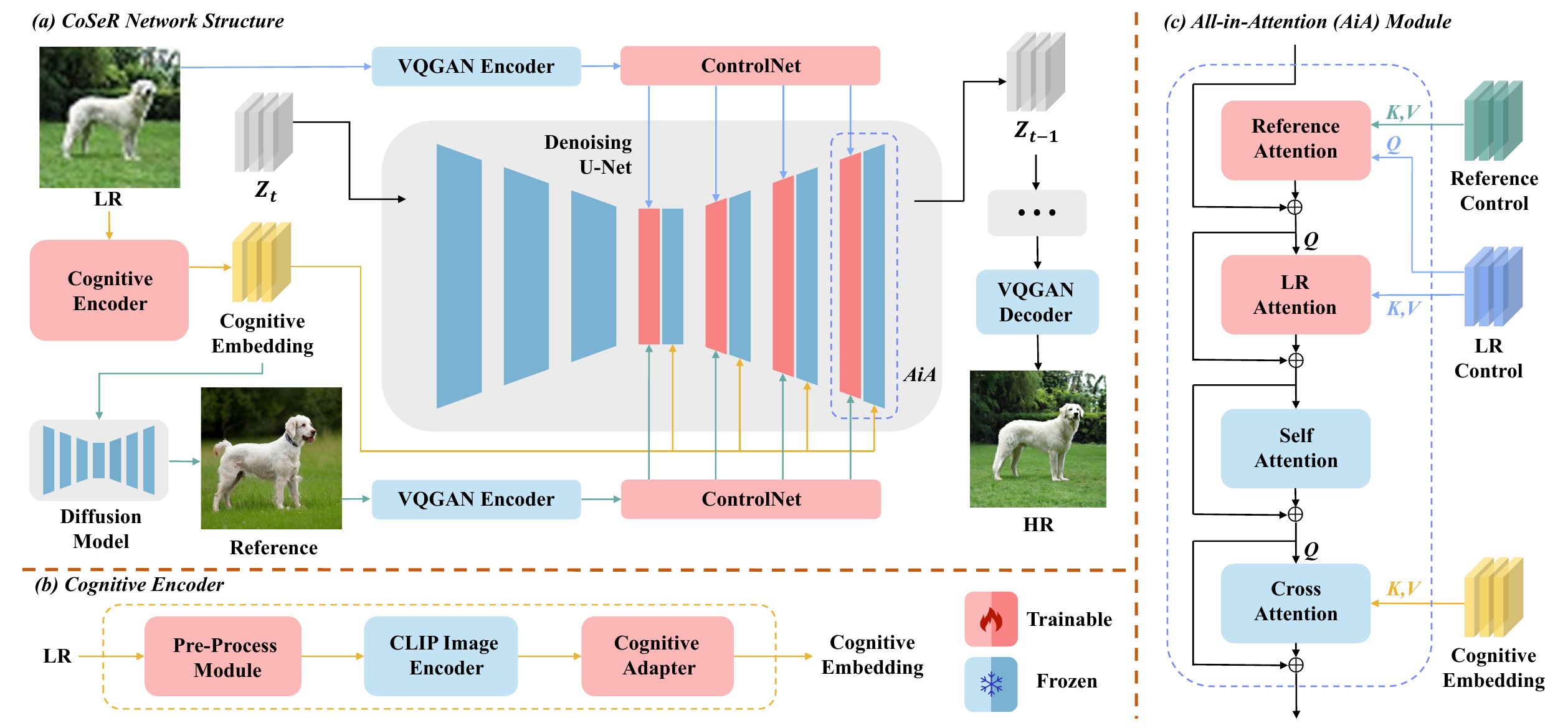}
    \vspace{-4mm}
    \caption{Framework of the proposed cognitive super-resolution (CoSeR) network. Given a low-resolution (LR) image, we employ a cognitive encoder to extract cognitive embedding containing semantic and textural information, which is then used to generate a high-quality reference image. The LR input, cognitive embedding, and reference image are integrated into the denoising U-Net using the all-in-attention (AiA) module, represented by blue, gold, and cyan lines, respectively. The structures of the cognitive encoder and AiA module are detailed in (b) and (c). Trainable modules are highlighted in red, while frozen modules are indicated in blue.}
    
    \label{fig:main}
    \vspace{-3mm}
\end{figure*}

\subsection{Reference-Based Super-Resolution}
%The reference graph serves as explicit prior information for the SR network and should ideally have content relevant to the low-resolution (LR) image to provide high-definition detail. Recent RefSR works can be divided into two branches~\cite{lu2021masa}: one branch focuses on spatial alignment, using methods like CrossNet~\cite{zheng2018crossnet} and SSEN~\cite{cao2022reference}, but these methods often struggle with long-distance correspondences. The other branch, represented by SRNTT ~\cite{zhang2019image} and TTSR~\cite{ttsr}, employs patch matching mechanisms for multi-level feature fusion. While this approach establishes long-range connections between the reference map and the LR image, it often comes at a higher computational cost. To address this issue, methods like MASA~\cite{lu2021masa} and CFE-PatchMatch~\cite{xia2022coarse} employ hierarchical correspondence matching and a coarse-to-fine matching strategy to improve computational efficiency and model robustness.

The reference image serves as an explicit prior, ideally containing content relevant to the LR image to facilitate the generation of high-definition details. Recent advancements in reference-based SR can be categorized into two branches~\cite{lu2021masa}. One branch prioritizes spatial alignment, employing techniques like CrossNet~\cite{zheng2018crossnet} and SSEN~\cite{cao2022reference}. However, these methods often encounter challenges in establishing long-distance correspondences. The other branch, represented by SRNTT~\cite{zhang2019image}, TTSR~\cite{ttsr}, MASA-SR~\cite{lu2021masa}, and CFE-PatchMatch~\cite{xia2022coarse}, utilizes patch-matching mechanisms to facilitate the establishment of long-range connections between the reference map and the LR image. 
% Building on this, approaches like MASA-SR~\cite{lu2021masa} and CFE-PatchMatch~\cite{xia2022coarse} use hierarchical correspondence matching to improve computational efficiency and model robustness. 
Yet, manually specifying reference images in real scenarios is labor-intensive, motivating the development of an automated and high-quality reference generation approach.

\section{Methodology}\label{sec:method}

Our Cognitive Super-Resolution (CoSeR) model employs a dual-stage process for restoring LR images. Initially, we develop a cognitive encoder to conduct a thorough analysis of the image content, conveying the cognitive embedding to the diffusion model. This enables the activation of pre-existing image priors within the pre-trained Stable Diffusion model~\cite{stablediffusion}, facilitating the restoration of intricate details. Additionally, our approach utilizes cognitive understanding to generate high-fidelity reference images that closely align with the input semantics. These reference images serve as auxiliary information, contributing to the enhancement of super-resolution results. Ultimately, our model simultaneously applies three conditional controls to the pre-trained Stable Diffusion model: the LR image, cognitive embedding, and reference image. The comprehensive framework is elucidated in Figure~\ref{fig:main}.

\subsection{Cognitive Encoder}
To distill cognitive information from LR images, our model commences with LR preprocessing aimed at mitigating the impact of degradation. Specifically, we employ a lightweight SRResnet~\cite{srgan} for 4$\times$ super-resolution, without additional supervision. Subsequently, we utilize a pre-trained CLIP~\cite{clip} image encoder to extract features from the preprocessed image. It is crucial to underscore that, although CLIP adeptly aligns image and language content, a significant disparity persists between the image embedding and the language embedding. These two components focus on different points, where image features inherently capture spatially variant details, while language features encapsulate comprehensive information. Consequently, a single language token may correspond to multiple subjects dispersed across diverse regions of an image.

\begin{figure*}[!htp]
    \centering
    \includegraphics[width=\textwidth]{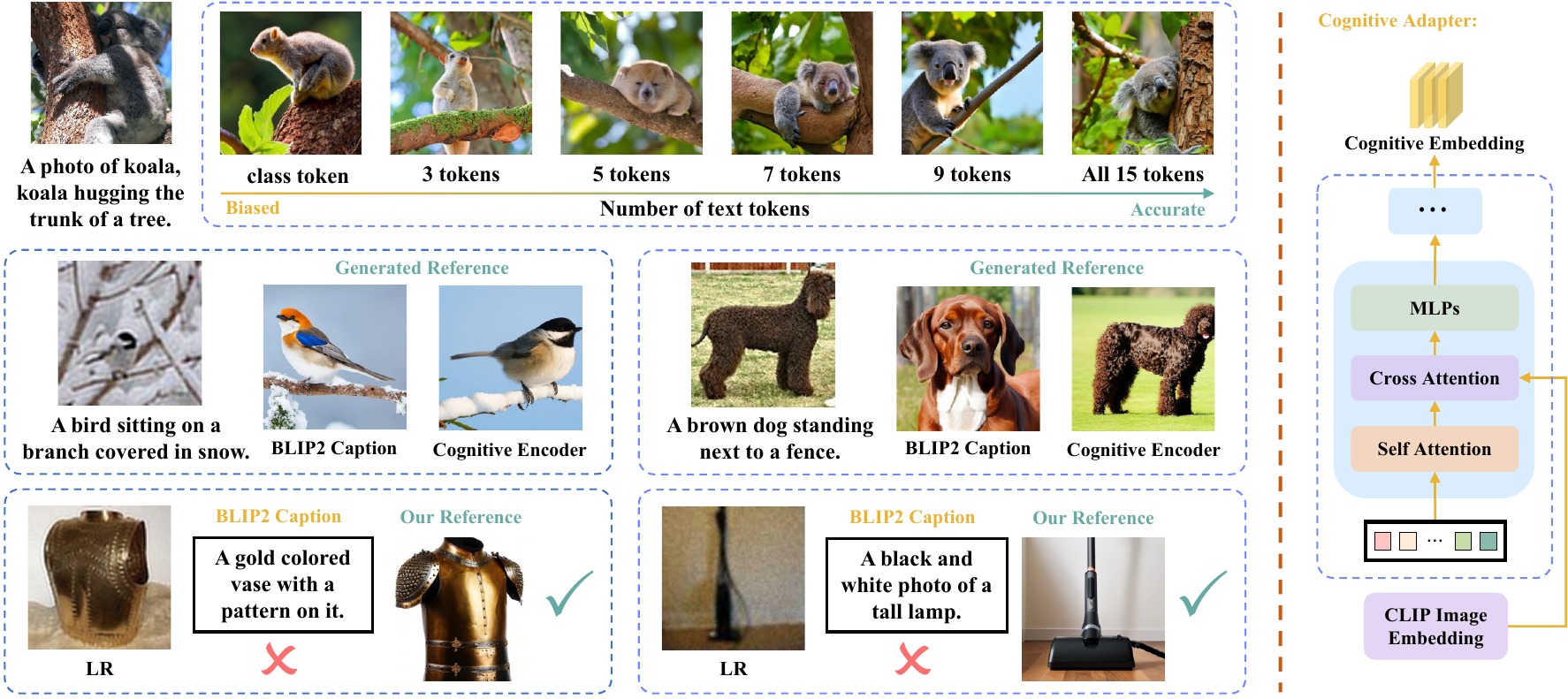}
    \caption{(\textbf{Left}) Generated reference images by BLIP2 captions and cognitive encoder. The first row shows the need to augment the token number. The last two rows show the drawbacks of directly employing captions for cognition. (\textbf{Right}) Structure of our cognitive adapter.}
    \label{fig:cognitive_encoder}
    \vspace{-3mm}
\end{figure*}

To overcome the challenge of aligning image and language representations, prior methods~\cite{ma2023unified, luo2023controlling} have often focused on aligning the class token of the image embedding and the class token of the corresponding language embedding, neglecting other tokens. However, relying solely on this single class token has been observed to introduce cognitive bias. As shown by the generated reference images from language tokens in the first row of Figure~\ref{fig:cognitive_encoder} (left part), cognitive bias diminishes gradually as the token number (before and including the class token) increases. To simultaneously address information misalignment and inaccurate cognition, we introduce a cognitive adapter that is tailored to extract multi-token cognitive embedding from image features, shown in Figure~\ref{fig:cognitive_encoder} ({right} part). Drawing inspiration from the Q-Former structure~\cite{blip2}, originally devised for vision-language representation learning, our approach employs learnable queries to interact with spatially-arranged image information, thereby reshaping information organization and facilitating feature compression. Our approach also incorporates a novel form of supervision, enhancing the adapter's capacity not only to reorganize image features but also to function as a modality transformer.

We represent the CLIP image embedding extracted from LR as $\bm{I} \in \mathbb{R}^{B \times T_i \times C_i}$, where $B, T_i, C_i$ denote batch size, token number, and channel number, respectively. Additionally, $\bm{L} \in \mathbb{R}^{B \times T_l \times C_l}$ denotes the CLIP language embedding extracted from the ground-truth caption (extracted from HR images using BLIP2~\cite{blip2}). In our cognitive adapter, we employ $T_e$ learnable queries ($T_e \leq T_l$) such that the resulting cognitive embedding is denoted as $\bm{E} \in \mathbb{R}^{B \times T_e \times C_l}$. We propose to use $T_e$ tokens preceding the class token $\bm{L}\left[t_{cls}\right]$ (inclusive) for supervision, as these tokens retain all previous information~\cite{clip}. If there are insufficient supervision tokens, we use the class token for end-filling. Therefore, our supervision $\bm{L^{\prime}}$ can be regarded as a more comprehensive representation than the class token. The loss function for training the cognitive encoder is expressed as:
\begin{equation}\label{cognitive_adapter_1}
\begin{aligned}
\mathcal{L}_{C E}=\left\|\bm{E}-\bm{L^{\prime}}\right\|_2^2, 
\end{aligned}
\end{equation}
where
\begin{equation}\label{cognitive_adapter_2}
\begin{aligned}
\bm{L^{\prime}}= \begin{cases}{\rm Padding}\left(\bm{L}\left[: t_{cls}\right], \bm{L}\left[t_{cls}\right]\right), & \text { if } t_{cls}<T_e ; \\ \bm{L}\left[\left(t_{cls}-T_e\right): t_{cls}\right], & \text { if } t_{cls} \geqslant T_e .\end{cases}
\end{aligned}
\end{equation}
A more extensive explanation of the supervision strategy can be found in the supplementary materials.

\noindent \textbf{Discussion.} We choose to utilize the feature embedding for the cognition process rather than directly generating a caption from LR for several compelling reasons. Firstly, although guided by language embedding, our cognitive embedding retains fine-grained image features, proving advantageous in generating reference images with high semantic similarity. In the second row of Figure~\ref{fig:cognitive_encoder} ({left} part), we show the BLIP2 captions generated from LR images. They fail to identify the precise taxon, color, and texture of the animals, leading to suboptimal generations compared to our cognitive adapter. Secondly, employing a pre-trained image caption model requires a substantial number of parameters, potentially reaching 7B~\cite{blip2}. In contrast, our cognitive adapter is significantly lighter, with only $3\%$ parameters, resulting in favorable efficiency. Thirdly, pre-trained image caption models may produce inaccurate captions for LR images due to disparities in the input distribution. In contrast, our cognitive adapter is more robust for LR images, shown in the third row of Figure~\ref{fig:cognitive_encoder} ({left} part).

\subsection{Reference Image Generation and Encoding}

We propagate the cognitive embedding to the pre-trained Stable Diffusion model for generating reference images without incurring additional parameters. The resulting reference images empower our SR model to explicitly leverage image priors. 
% As depicted in Figure~\ref{fig:teaser}, our cognitive embedding excels in producing well-aligned reference images, adept at capturing fine-grained semantics, such as precise animal and plant categories.
As depicted in Figure~\ref{fig:teaser}, our cognitive embedding excels in producing well-aligned reference images.

We employ a pre-trained VQGAN~\cite{vqgan} for encoding images into latent codes, as opposed to a trainable CNN like~\cite{controlnet}, given the robust encoding capabilities exhibited by VQGAN. Subsequently, we follow the ControlNet~\cite{controlnet} approach by utilizing the U-Net encoder to generate multi-scale control features. We represent the LR control and reference image control as $\{\bm{X}_i\}^{4}_{i=1}$ and $\{\bm{R}_i\}^{4}_{i=1}$, respectively. Notably, we observe that using a single control encoder for both LR and reference images is sufficient for achieving satisfactory results, enhancing the parameter efficiency of our model. The generated controls are then input into the All-in-Attention module, as elaborated in the following section. In fact, when automatically generating reference images, we only need to use Stable Diffusion to generate latent codes, and subsequently input them into the control encoder, thereby circumventing the process of decoding and encoding in Figure~\ref{fig:main}. 
% However, the encoding process is necessary if the user wants to use a customized reference image.

\subsection{All-in-Attention Module}
In image super-resolution, preserving fidelity to LR inputs is important. Our experiments in Section~\ref{sec:abl_aia} demonstrate that the introduction of LR control $\{\bm{X}_i\}^{4}_{i=1}$ through attention mechanisms leads to enhanced fidelity. Consequently, we advocate for the comprehensive integration of all conditional information into our model, achieved through the design of an All-in-Attention (AiA) module. Beyond accommodating LR inputs, this design facilitates reference patch-matching for the establishment of long-range connections~\cite{ttsr}. The cognitive embedding is seamlessly incorporated via the cross-attention mechanism of Stable Diffusion. 

Illustrated in Figure~\ref{fig:main} (c), the AiA module enhances the original attention module in Stable Diffusion by introducing trainable reference attention and LR attention, while maintaining the frozen state of the self-attention and cross-attention components. This structural augmentation is applied across all attention modules within the middle and decoder of the denoising U-Net. We denote the query, key, and value features in the attention mechanism as $\bm{Q}$, $\bm{K}$, and $\bm{V}$, respectively. Regarding LR attention, $\bm{Q}$ is derived from the denoising U-Net feature $\bm{Z}$, while $\bm{K}$ and $\bm{V}$ originate from the LR control $\bm{X}_i$. In reference attention, we opt to use the LR control as $\bm{Q}$ for better fidelity, with $\bm{K}$ and $\bm{V}$ coming from the reference control $\bm{R}_i$. In the original cross-attention, we use cognitive embedding $\bm{E}$ as inputs for $\bm{K}$ and $\bm{V}$. 
% Additionally, to prevent the newly introduced attention components from influencing the well-established representation in Stable Diffusion during early training, we integrate zero convolutions~\cite{controlnet} at the end of the reference attention and LR attention. 
Notably, to counteract the potential blurring effect of the conventional attention mechanism in reference-based SR~\cite{ttsr}, we introduce ``one-hot attention'' to enhance the LR image with the most relevant reference feature, and additional details are available in the supplementary materials.

\begin{table*}[htp]
\begin{center}

\scalebox{0.89}{
\begin{tabular}{c|c|ccccccc|c}
\toprule
Datasets                                                                      & Metrics            & RealSR & Real-ESRGAN+ & BSRGAN  & DASR   & FeMaSR        & LDM    & StableSR & CoSeR (Ours)          \\ \hline\hline
\multirow{6}{*}{\begin{tabular}[c]{@{}c@{}}ImageNet \\ Test2000\end{tabular}} & FID$\downarrow$      &      86.36                      & 32.68        & 41.11  & 39.15  & 31.25          & 34.54  & \underline{22.53}    & \textbf{19.41}  \\
                                                                              & DISTS$\downarrow$    &     0.2649                       & 0.1739       & 0.1946 & 0.1931 & 0.1597         & 0.1664 & \underline{0.1527}   & \textbf{0.1482} \\
                                                                              & LPIPS$\downarrow$    &        0.4519                    & 0.2943       & 0.3381 & 0.3346 & 0.3027         & 0.3289 & \underline{0.2871}   & \textbf{0.2863} \\
                                                                              & CLIP-Score$\uparrow$ &        0.6242                    & 0.8132       & 0.7719 & 0.7838 & 0.8253         & 0.8119 & \underline{0.8622}   & \textbf{0.8755} \\
                                                                              & MANIQA$\uparrow$     &        0.0796                    & 0.1370       & 0.1115 & 0.0914 & \underline{0.1936}         & 0.1830 & 0.1556   & \textbf{0.2133} \\
                                                                              & MUSIQ$\uparrow$      &        50.18                    & 57.52        & 52.33  & 48.98  & \textbf{67.20} & 64.15  & 60.20    & \underline{65.51}           \\ \midrule
\multirow{6}{*}{RealSR~\cite{realsr}}                                                       & FID$\downarrow$      &          157.85                  & 87.00        & 111.03 & 107.38 & 91.45          & 92.43  & \underline{84.06}    & \textbf{80.82}  \\
                                                                              & DISTS$\downarrow$    &        0.2529                    & 0.2028       & 0.2545 & 0.2171 & 0.2131         & 0.2055 & \underline{0.1867}   & \textbf{0.1826} \\
                                                                              & LPIPS$\downarrow$    &        0.3672                    & 0.2803       & 0.3224 & 0.3056 & 0.2683         & 0.2924 & \underline{0.2536}   & \textbf{0.2438} \\
                                                                              & CLIP-Score$\uparrow$ &        0.7458                    & 0.8345       & 0.8074 & 0.8332 & 0.8108         & 0.8330 & \underline{0.8517}   & \textbf{0.8545} \\
                                                                              & MANIQA$\uparrow$     &        0.1474                    & 0.1776       & 0.1696 & 0.1803 & 0.2033         & 0.1986 & \underline{0.2144}   & \textbf{0.2522} \\
                                                                              & MUSIQ$\uparrow$      &        60.40                    & 61.90        & 60.82  & 60.90  & 66.47          & \underline{67.27}  & 67.08    & \textbf{70.29}  \\ \midrule
\multirow{6}{*}{DRealSR~\cite{drealsr}}                                                      & FID$\downarrow$      &          148.58                  & \underline{74.72}        & 107.76 & 96.42  & 86.81          & 87.16  & 75.83    & \textbf{71.22}  \\
                                                                              & DISTS$\downarrow$    &        0.2673                    & 0.2216       & 0.2238 & 0.2345 & 0.2231         & 0.2179 & \underline{0.2048}   & \textbf{0.1977} \\
                                                                              & LPIPS$\downarrow$    &             0.4212               & 0.3239       & 0.3972 & 0.3534 & 0.2981         & 0.3258 & \underline{0.2920}   & \textbf{0.2702} \\
                                                                              & CLIP-Score$\uparrow$ &             0.7360               & 0.8504       & 0.8157 & 0.8510 & 0.8332         & 0.8459 & \underline{0.8681}   & \textbf{0.8766} \\
                                                                              & MANIQA$\uparrow$     &             0.1090               & 0.1742       & 0.1491 & 0.1739 & 0.1998         & 0.1890 & \underline{0.2241}   & \textbf{0.2575} \\
                                                                              & MUSIQ$\uparrow$      &        54.28                    & 62.80        & 57.72  & 62.14  & 66.57          & 67.03  & \underline{68.27}    & \textbf{70.18}  \\ \bottomrule
\end{tabular}
}
\vspace{-1mm}
\caption{Quantitative comparisons on both ImageNet Test2000 and real-world benchmarks (RealSR and DRealSR). 
% The best results are highlighted in \textbf{bold}.
The best results are highlighted in \textbf{bold} and the second best results are in \underline{underlined}.
\vspace{-7mm}
}\label{tab:main_comparison}
\end{center}

\end{table*}

\section{Experiments}

\subsection{Implementation Details}

Our CoSeR is built based on Stable Diffusion 2.1-base\footnote{https://huggingface.co/stabilityai/stable-diffusion-2-1-base}. The model is trained with a batch size of 192 over 20000 steps on 8 V100 GPUs. We use Adam~\cite{adam} optimizer with a learning rate of $5\times10^{-5}$. Following StableSR~\cite{stablesr}, we train our model on $512\times512$ resolution and apply DDPM sampling~\cite{ddpm} with 200 timesteps for inference.

% Our training comprises two stages: initially, we train the cognitive encoder independently using the defined loss function Eq.~\ref{cognitive_adapter_1}. Subsequently, we freeze the cognitive encoder and proceed to train the complete model. The number of learnable queries in our cognitive encoder is experimentally set to $50$, with detailed justifications provided in the supporting materials. Following~\cite{controlnet}, we initialize ControlNet with Stable Diffusion weights. To maximize the utilization of the pre-trained model, we initialize reference attention and LR attention with self-attention weights within each all-in-attention module. During training, the user-provided prompt embedding has a $50\%$ probability of using the ground-truth caption text embedding and a $50\%$ probability of using a null text embedding. For inference, we enhance the guidance effect using the classifier-free guidance method~\cite{classifier-free}, with the guidance scale fixed at $3$. We also combine words like ``low quality'', ``blurry'', ``low resolution'', \textit{etc.}, as negative prompts. To achieve a flexible trade-off between realism and fidelity, we adopt the pre-trained controllable feature wrapping~\cite{stablesr}.
The training process involves two stages: Firstly, we train the cognitive encoder using the defined loss function in Eq.~\ref{cognitive_adapter_1}. The cognitive encoder employs 50 learnable queries, a choice substantiated in the supplementary materials. Then, we freeze the cognitive encoder and train the SR model. 
% We illustrate more detailed modeling details and training strategies in the supplementary material. 
Following~\cite{controlnet}, we initialize ControlNet with Stable Diffusion weights. To maximize the utilization of the pre-trained model, the reference attention and LR attention modules are initialized using self-attention weights. 
% During training, the user-provided prompt is randomly assigned a 50$\%$ probability of using either the ground-truth caption or null text. In the inference phase, if no user-provided prompt is available, we use null text as the prompt. 
In the inference phase, cognitive information is enhanced via classifier-free guidance~\cite{classifier-free}, utilizing a scaling factor of 3. To optimize the trade-off between realism and fidelity, we adopt the pre-trained feature wrapping module in~\cite{stablesr}, which is integrated with the VQGAN decoder.

\begin{figure*}[htp]
    \centering
    \includegraphics[width=\textwidth]{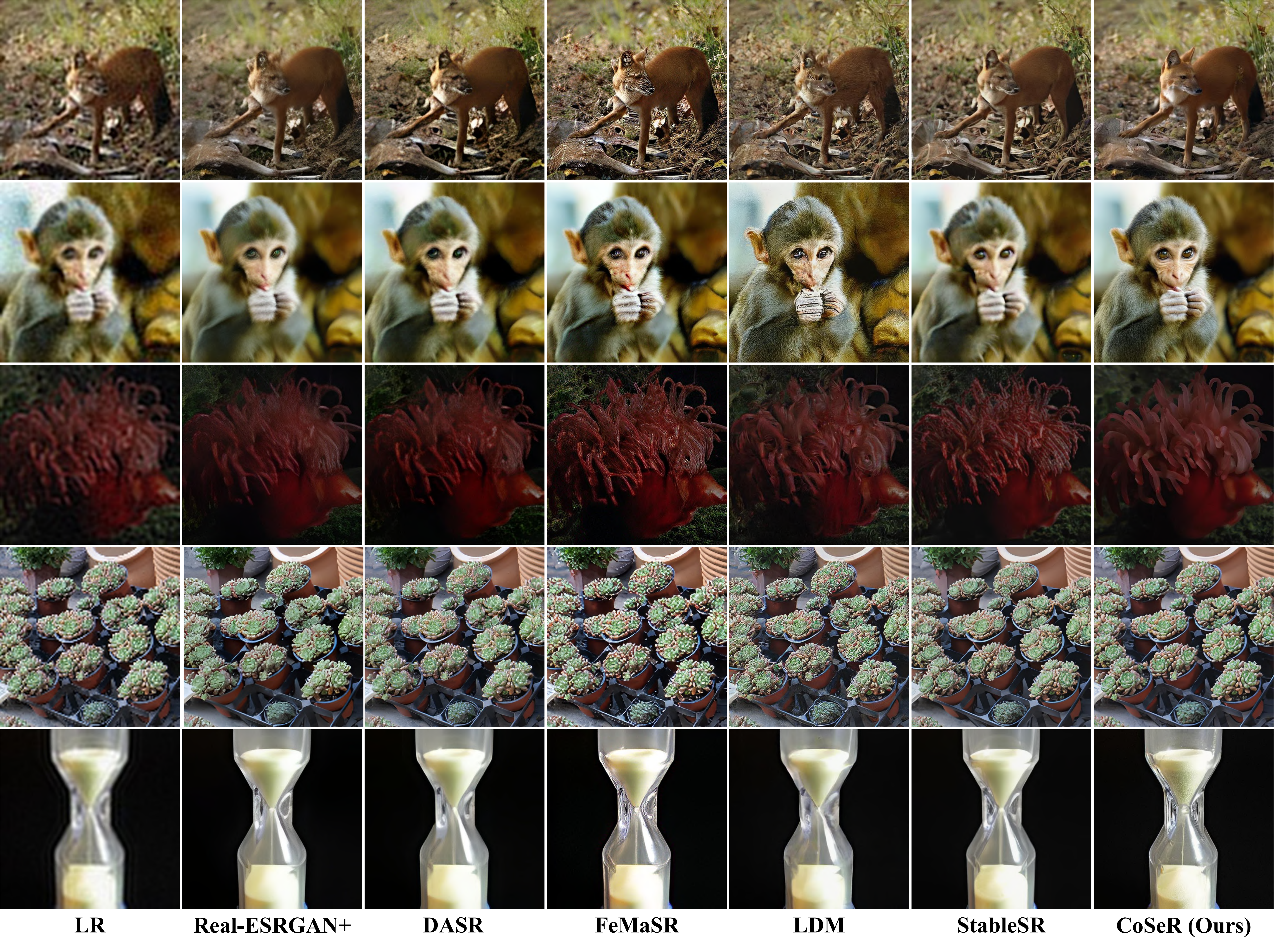}
    \vspace{-7mm}
    \caption{Qualitative comparisons on both synthesized and real-world test datasets. Our CoSeR obtains the best visual performance.
    % The first six lines are from ImageNet Test2000 dataset and the last second lines are from RealSR and DRealSR datasets, respectively.
    }
    \vspace{-2mm}
    \label{fig:main_comparison}
\end{figure*}

\subsection{Experimental Settings}
\noindent \textbf{Training and testing datasets.} 
% Traditional super-resolution models are usually trained on datasets such as DIV2K~\cite{div2k}, Flickr2K~\cite{flickr2k} and OutdoorSceneTraining~\cite{ost}. The images in these datasets usually have very high resolution, and during training, we need to crop them into smaller patches to feed into the network. However, our CoSeR needs to receive full-scene perception of input images, this approach is not suitable for us. 
We aim to develop an image super-resolution model empowered with cognitive capabilities adaptable to diverse real-world scenarios. To this end, we utilize the extensive ImageNet dataset~\cite{imagenet} for training, renowned for its wide array of scenarios and objects. We acquire over 900K HR images with $512\times512$ resolution and employ Real-ESRGAN~\cite{real-esrgan} degradation to generate corresponding LR images. We employ BLIP2~\cite{blip2} to generate three descriptive captions for each HR image, filtering out captions with CLIP scores~\cite{clip} below $0.28$. 
% For training the reference-based SR module, we identify the ten most similar images within each category and randomly select one as the reference image.

To comprehensively assess our model's performance across diverse scenarios, we curate a non-overlapped ImageNet test set consisting of 2000 LR-HR pairs using the Real-ESRGAN pipeline. We choose two images from each category, ensuring the test set's diversity and balance. In addition to our constructed test set, we conduct evaluations on established real-world benchmarks such as RealSR~\cite{realsr} and DRealSR~\cite{drealsr}. In this section, LR images are acquired at the same resolution used during training, specifically $128\times128$. For datasets such as RealSR and DRealSR, we initially resize LR images, adjusting the shorter sides to $128$, followed by center cropping.

\noindent \textbf{Compared methods.} 
% We compare our CoSeR with several state-of-the-art real-world SR methods, including Real-ESRGAN+~\cite{real-esrgan}, SwinIR-GAN~\cite{swinir}, BSRGAN~\cite{bsrgan}, FeMaSR~\cite{femasr}, latent diffusion model (LDM)\cite{stablediffusion}, StableSR\cite{stablesr}. For the sake of fairness in comparison, we train the diffusion based model LDM and StableSR with the same training settings as our CoSeR. For other methods, we follow the official training settings.
We compare CoSeR with some state-of-the-art real-world SR methods, including RealSR~\cite{realsr_method}, Real-ESRGAN+~\cite{real-esrgan}, BSRGAN~\cite{bsrgan}, DASR~\cite{dasr}, FeMaSR~\cite{femasr}, latent diffusion models (LDM)~\cite{stablediffusion}, StableSR~\cite{stablesr}. To ensure fair comparisons, we retrain all these models using our ImageNet training set except RealSR and BSRGAN, which share the network structure with Real-ESRGAN+ but employ different degradation pipelines.

\noindent \textbf{Evaluation metrics.} 
% We use five perceptual metrics to evaluate the quality of SR results, including FID~\cite{fid}, DISTS~\cite{dists}, LPIPS~\cite{lpips}, CLIP-Score~\cite{clip}, and MUSIQ~\cite{musiq}. FID, DISTS, and LPIPS were used to gauge the perceptual distance between SR outcomes and high-resolution (HR) images. CLIP-Score measured the semantic accuracy of the results by calculating scores between HR images and SR results. Since our approach is oriented towards real-world scenarios where the ground-truth HR is not available, we also incorporated a non-reference image quality assessment MUSIQ. Pixel-level image quality assessments such as PSNR and SSIM are reported in the supplementary materials, as prior works~\cite{dists, lpips, maniqa} have highlighted their limitations in aligning with human perception.
To better align with human perception, we employ six perceptual metrics: FID~\cite{fid}, DISTS~\cite{dists}, LPIPS~\cite{lpips}, CLIP-Score~\cite{clip}, MANIQA~\cite{maniqa} and MUSIQ~\cite{musiq}. FID, DISTS, and LPIPS measure perceptual distance, while CLIP-Score estimates semantic accuracy by evaluating scores between HR images and SR results. Given our focus on real-world scenarios where ground-truth HR data might be unavailable, we include non-reference image quality assessments, MANIQA and MUSIQ. Notably, pixel-level image quality assessments like PSNR and SSIM are presented in the supplementary materials solely for reference. Prior research~\cite{dists, lpips, maniqa, shi2021region} has indicated their weak correlation with human perception regarding image quality in real-world contexts.
% For the comparison experiments, we did not use pixel-level image quality assessments such as PSNR and SSIM, as many existing works~\cite{dists, lpips, maniqa} have pointed out that they do not match well with human perception. However, we will use them when evaluating certain specific aspects of SR results, such as fidelity.

\subsection{Comparison with State of the Arts}
\noindent \textbf{Quantitative comparison.} 
% We conducted a quantitative comparison of our method with other real-world image super-resolution (SR) techniques on both generated (ImageNet Test2000) and real datasets (RealSR and DRealSR). To ensure a fair comparison on ImageNet Test2000, previous methods were retrained on our ImageNet training dataset, while for common real-world benchmarks, we prioritized using the official training models. Table~\ref{tab:main_comparison} presents the results, indicating that our method consistently achieves top performance across almost all datasets and metrics, showcasing its superiority and robustness. Specifically, the FID scores of our method are $\bf{13.\%}$, $\bf{12.6\%}$, and $\bf{14.6\%}$ lower than the second-best performance on ImageNet Test2000, RealSR, and DRealSR, respectively. Although FeMaSR outperforms our method in MUSIQ on the ImageNet Test2000 dataset, as illustrated in Fig.~\ref{fig:main_comparison}, it introduces numerous unrealistic artifacts that may impact the accuracy of non-reference assessments.
We perform an extensive quantitative comparison on both the ImageNet Test2000 dataset and real-world benchmarks (RealSR and DRealSR), as presented in Table~\ref{tab:main_comparison}. As mentioned previously, we retrain the comparison models using the ImageNet training dataset to ensure fair comparisons. Additionally, the results of officially released models are provided in the supplementary materials. Our method consistently demonstrates superior performance across nearly all datasets and metrics, highlighting its robustness and superiority. Notably, our FID scores surpass the second-best performance by $\bf{13.8\%}$, $\bf{3.8\%}$, and $\bf{4.7\%}$ on the ImageNet Test2000, RealSR, and DRealSR, respectively. While FeMaSR exhibits better performance in MUSIQ on the ImageNet Test2000, as depicted in Figure~\ref{fig:main_comparison}, it introduces numerous unrealistic artifacts that might not be reflected by the non-reference metric MUSIQ.

\begin{figure*}[htp]
    \centering
    \includegraphics[width=\textwidth]{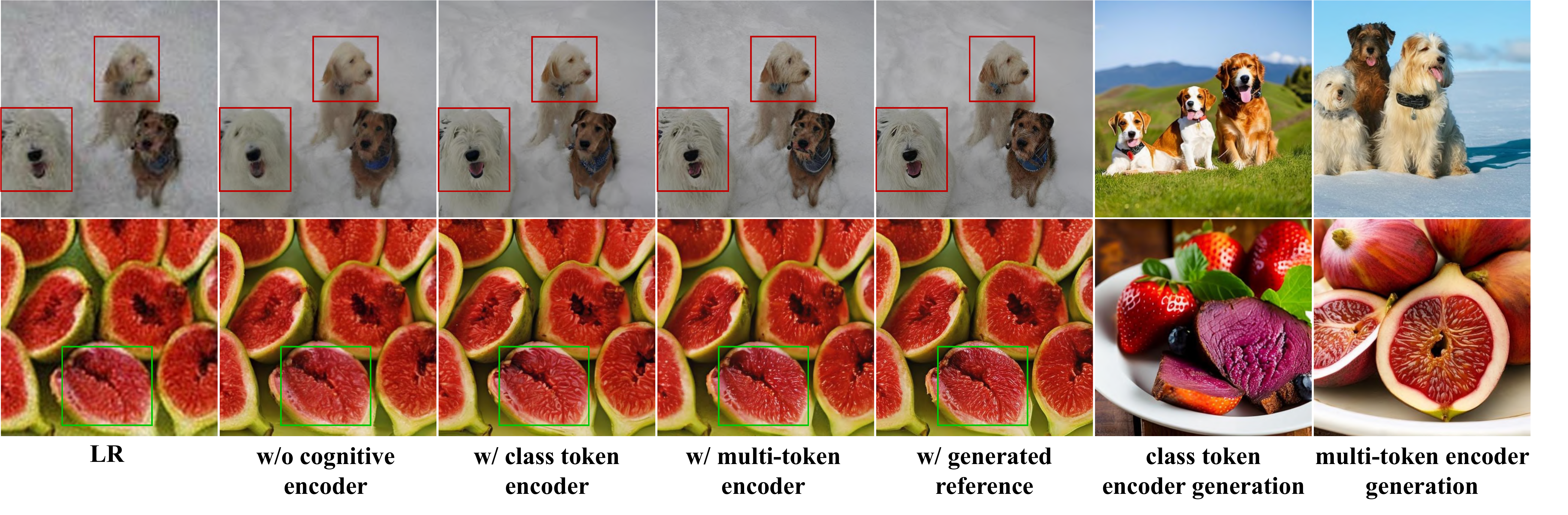}
    \vspace{-6mm}
    \caption{Visual comparisons were conducted to assess the impact of cognitive information, comparing scenarios with no utilization of cognitive information, employing different cognitive encoders, and incorporating the generated reference image.}
    \vspace{-2mm}
    \label{fig:abl}
\end{figure*}

\noindent \textbf{Qualitative comparison.} 
% In Fig.~\ref{fig:main_comparison}, we present visual comparisons between our method and others. Firstly, our approach excels in generating high-quality texture details for severely blurred input images. Illustrated in the first, third, and fourth rows, our results showcase more realistic and clearer fur and facial features in animals. Secondly, leveraging cognitive abilities empowers our method to comprehend scene information, enabling the introduction of appropriate textures. Evident in the second, fifth, and last rows, our method accurately recovers textures such as anemone tentacles, cable knots, and succulent leaves, a feat unmatched by other methods. Lastly, the cognitive capability enables our model to recover semantically indispensable details that are nearly lost in low-resolution (LR). For instance, in the third row, only our model successfully recovers the dhole's eyes; in the sixth row, our method uniquely restores the sand in the hourglass.
We provide visual comparisons in Figure~\ref{fig:main_comparison}. Enriched by a comprehensive understanding of scene information, CoSeR excels in enhancing high-quality texture details. As demonstrated in the first and second rows, our results exhibit significantly clearer and more realistic fur and facial features in the animals. Similarly, in the third and fourth rows, our method adeptly reconstructs realistic textures such as the anemone tentacles and succulent leaves—achievements unmatched by other methods. Particularly, our model's cognitive capabilities enable the recovery of semantic details almost lost in low-resolution inputs. Notably, in the first row, only our model successfully restores the dhole's eyes, while in the fifth row, only our method can reconstruct the sand within the hourglass. These visual cases distinctly showcase our model's capacity to comprehend scenes and produce high-quality images.

\noindent\textbf{User Study.} 
To further substantiate the effectiveness of CoSeR in real-world scenarios, a user study is conducted on 20 real-world LRs collected from the Internet or captured by mobile phones. 23 subjects are asked to select the visually superior result from the four HRs generated by Real-ESRGAN+, FeMaSR, StableSR, and CoSeR. A total of $20 \times 23$ votes are collected, with approximately $\bf{80\%}$ of participants concurring that our method exhibited the best visual effect. This underscores the superiority and robustness of CoSeR in real-world scenarios. Detailed voting results are available in the supplementary materials.

\subsection{Ablation Study}
We dissect the individual contributions of different components in our framework. Given the diverse focuses of different components, we employ the most appropriate evaluation metrics to measure their respective utilities.

\noindent\textbf{Cognitive information.} 
% As shown in Table~\ref{tab:abl}, with the use of cognitive information, our method shows a significant improvement in FID and CLIP-Score (for fairness, we kept the negative prompts and classifier-free guidance for the situation without conition). Fig.~\ref{fig:abl} shows that cognitive information helps restore severely degraded details, such as animal hair and pulp texture. Following~\cite{luo2023controlling}, we implement the class token adapter with an MLP. Our results show that merely using one class token to convey cognitive information is prone to cognitive bias, which can be illustrated by the generated reference images. Here we compute the CLIP-Score between the generated reference images and the HR images as a new metric, namely ``Gen-score''. It can be seen that the generation quality of our cognitive adapter is significantly better than the class token approach. Our Advantages are also reflected in the SR results. As can be seen in the pulp texture in Fig.~\ref{fig:abl}, the introduced textures by our adapter are more semantically accurate.
% In Table~\ref{tab:abl}, the incorporation of cognitive information in our method demonstrates a notable enhancement in both FID and CLIP-Score. Notably, we maintained negative prompts and applied classifier-free guidance for the \textit{w.o. cognition} condition to ensure fairness. The visual representation in Fig.~\ref{fig:abl} reinforces that cognitive information contributes substantially to the restoration of severely degraded details, such as animal hair and pulp texture.
Prior research, such as~\cite{luo2023controlling}, has attempted to align class tokens between CLIP image and text encoders using MLP. In our terminology, we denote the cognitive encoder with class token alignment MLP and our cognitive encoder as the ``class token encoder'' and ``multi-token encoder'', respectively. As shown in Table~\ref{tab:abl}, integrating cognitive information significantly enhances FID and CLIP-Score metrics, signifying a more accurate generation of semantics and textures. Our investigations reveal that the class token encoder may introduce semantic and texture biases, as evident in the quality of the generated reference images in Figure~\ref{fig:abl} (penultimate column). To quantitatively evaluate cognitive bias, we introduce a new metric termed ``Gen-score", calculated as the CLIP-Score between the generated reference image and the ground-truth image. Both the metrics and the visuals distinctly highlight our superior cognitive ability. This advantage extends to the final SR results, notably visible in the precise introduction of the hair and pulp texture in Figure~\ref{fig:abl}.

\noindent\textbf{Reference guidance.} 
% The explicitly introduced reference image helps to further enhance the texture details of the SR results (Fig.~\ref{fig:abl}), improving the subjective perception of the human eye. Therefore, here we mainly show the effect through FID and two non-reference image quality assessments, MUSIQ and MANIQA~\cite{maniqa}. As shown in Table.~\ref{tab:abl}, the generated reference image improves the overall visual quality of the SR results without compromising the fidelity of the results. Besides, compared to using the most similar image in ImageNet as the reference, our generated image can achieve comparable or even superior results. 
The explicitly introduced reference image significantly contributes to enhancing the texture details in the SR results (Figure~\ref{fig:abl}). To better correlate with human perception, our evaluation primarily focuses on assessing restoration quality using FID and two non-reference image quality assessments. As demonstrated in the fifth column of Table.~\ref{tab:abl}, the inclusion of the generated reference image notably elevates the overall visual quality of the SR results without compromising their fidelity. Additionally, when compared to the utilization of real-world reference images from ImageNet, our generated image achieves comparable or even better results.

\begin{table}[!t]
\begin{center}
\scalebox{0.9}{
\begin{tabular}{c|ccc}
\toprule
\rowcolor[gray]{0.9}
CoSeR                  & FID$\downarrow$ & CLIP-Score$\uparrow$ & Gen-score$\uparrow$ \\ \midrule
w/ multi-token encoder   & \textbf{20.27}  & \textbf{0.8674}      & \textbf{0.5953}          \\
w/ class token encoder & 21.35           & 0.8628               & 0.4881                   \\
w/o cognitive encoder         & 23.18           & 0.8484               & $-$                      \\ \midrule
\rowcolor[gray]{0.9}
CoSeR                  & FID$\downarrow$ & MUSIQ$\uparrow$      & MANIQA$\uparrow$         \\ \midrule
w/ generated reference & 19.80           & \textbf{64.21}       & \textbf{0.2107}          \\
w/ ImageNet reference  & \textbf{19.72}  & 63.49                & 0.2056                   \\
w/o reference         & 20.27           & 61.82                & 0.1874                   \\ \midrule
\rowcolor[gray]{0.9} 
CoSeR                  & FID$\downarrow$ & DISTS$\downarrow$    & LPIPS$\downarrow$        \\ \midrule
w/ AiA                 & \textbf{20.27}  & \textbf{0.1502}      & \textbf{0.3076}          \\
w/ SFT                 & 21.50           & 0.1530               & 0.3101    \\ \bottomrule
\end{tabular}
}
\vspace{-1mm}
\caption{Ablation studies on cognitive information, reference guidance, and All-in-Attention (AiA) module on ImageNet Test2000. We remove controllable feature wrapping~\cite{stablesr} for evaluation.}\label{tab:abl}

\vspace{-9mm}
\end{center}
\end{table}

\noindent\textbf{All-in-Attention (AiA) module.} \label{sec:abl_aia}
% We utilize FID, DISTS, and LPIPS metrics to assess the impact of our All-in-Attention (AiA) module on enhancing the fidelity of super-resolution (SR) results. In comparison with the spatial feature transformations (SFT) method~\cite{sftgan} implemented in StableSR~\cite{stablesr}, our AiA module achieves a FID score that is $5.7\%$ lower, underscoring its effectiveness in enhancing result fidelity.
Excessive generation sometimes results in a compromise in fidelity. To address this, we introduced the All-in-Attention (AiA) module designed to incorporate multiple conditions, aiming to enhance consistency with the input image. To evaluate fidelity, we utilize ground-truth-involved FID, DISTS, and LPIPS metrics. Compared to spatial feature transform (SFT)~\cite{sftgan} integrated in StableSR~\cite{stablesr}, our AiA module achieves a $5.7\%$ lower FID score, along with superior DISTS and LPIPS results. This manifests the effectiveness of our AiA module in enhancing result fidelity.

\section{Conclusion}
In this paper, we present a pioneering approach to endow super-resolution (SR) with cognitive abilities. Our model excels in producing high-definition reference images that aid the SR process. Furthermore, we introduce an All-in-Attention module to enhance result fidelity. Extensive experiments substantiate the effectiveness of our approach in real-world applications.

\clearpage

% \bf{Supplementary}
\begin{center}
    \Large
    % Supplementary
    \bf{Supplementary Material}
\end{center}

\setcounter{section}{1}

\renewcommand\thesection{\Alph{section}}
\renewcommand\thesubsection{\thesection.\arabic{subsection}}
\renewcommand\thefigure{\Alph{section}.\arabic{figure}}
\renewcommand\thetable{\Alph{section}.\arabic{table}} 

%\begin{abstract}
Sec.~\ref{sec1} provides an extensive elucidation of our method, including details of the cognitive encoder supervision method, the one-hot reference attention mechanism, and an in-depth analysis of the network architectures governing the denoising U-Net and ControlNet. In Sec.~\ref{sec2}, we present comprehensive quantitative comparisons between the proposed method and established models, including the recently introduced DiffBIR~\cite{diffbir}. Additionally, our investigation delves into the impact of introducing multiple generated reference images. We provide the results of a user study in the form of voting results and assessments of image quality at the pixel level. Sec~\ref{sec3} showcases more visualization examples, extensively demonstrating the effectiveness of our method. Finally, we talk about the future work in~\ref{sec4}.

%We first present more detailed explanations of our method in Sec.~\ref{sec1}, including the cognitive encoder supervision method, one-hot reference attention, and network structures of the denoising U-Net and ControlNet. Sec.\ref{sec2} illustrates quantitative comparisons with officially released models, including the newly added DiffBIR~\cite{diffbir}. Additionally, we explore the effect of introducing multiple generated reference images. For a more comprehensive analysis, we also show the voting results of the user study and pixel-level image quality assessments. Finally, we provide more visualization examples in Sec.~\ref{sec3}. 

% We achieve this by marrying specific image features and generalized language understanding to generate a cognitive embedding for comprehensive scene cognition. This embedding not only extracts prior information from large text-to-image diffusion models but also facilitates the generation of high-quality reference images to optimize the SR process.
%\end{abstract}

\section{Detailed Illustration of our Method}\label{sec1}
\subsection{Cognitive Encoder Supervision}
As aforementioned in the main paper, we use $T_e$ ($T_e \leq T_l$) tokens, preceding the class token $\bm{L}\left[t_{cls}\right]$ (inclusive), extracted from the CLIP language embedding $\bm{L} \in \mathbb{R}^{B \times T_l \times C_l}$ for supervision. $B, T_l$, and $C_l$ denote batch size, token number, and channel number, respectively. If there are insufficient supervision tokens, we use the class token for end-filling. The loss function for training the cognitive encoder is expressed as:
\begin{equation}\label{cognitive_adapter_1}
\begin{aligned}
\mathcal{L}_{C E}=\left\|\bm{E}-\bm{L^{\prime}}\right\|_2^2, 
\end{aligned}
\end{equation}
where
\begin{equation}\label{cognitive_adapter_2}
\begin{aligned}
\bm{L^{\prime}}= \begin{cases}{\rm Padding}\left(\bm{L}\left[: t_{cls}\right], \bm{L}\left[t_{cls}\right]\right), & \text { if } t_{cls}<T_e ; \\ \bm{L}\left[\left(t_{cls}-T_e\right): t_{cls}\right], & \text { if } t_{cls} \geqslant T_e .\end{cases}
\end{aligned}
\end{equation}

% We observed that directly using $\bm{L}$ as the supervision for cognitive embedding $\bm{E}$ (setting $T_e = T_l$) hinders the learning of cognitive information. As shown in Fig.~\ref{fig:supervision}, the generated reference images have no relation to LR. This is attributed to the varying length of captions, causing the learnable queries in the Q-Former to inadequately capture semantic information at corresponding positions. To address this, we propose utilizing the last-$T_e$ tokens for supervision. Two rationales support this choice: Firstly, leveraging the "mnemonic" nature of CLIP text encoder encoding, backward tokens retain the memory of all preceding words~\cite{clip}. Consequently, our use of last-$T_e$ tokens weakens the correlation between semantic representations and the sequential ordering of words, aiding easier learning by the queries. Secondly, in our supervision $\bm{L^{\prime}}$, the last query is consistently aligned with the class token, thereby preserving the original representational power of the class token. Our last-$T_e$ supervision can be considered as a more comprehensive overall representation than the class token. Both Fig. 5 in the main text and Fig.~\ref{fig:supervision} demonstrate that our proposed supervised approach to learning cognitive encoder has better cognitive capabilities than supervising with a single class token.

\begin{figure}[!tp]
    \centering
    \includegraphics[width=0.49\textwidth]{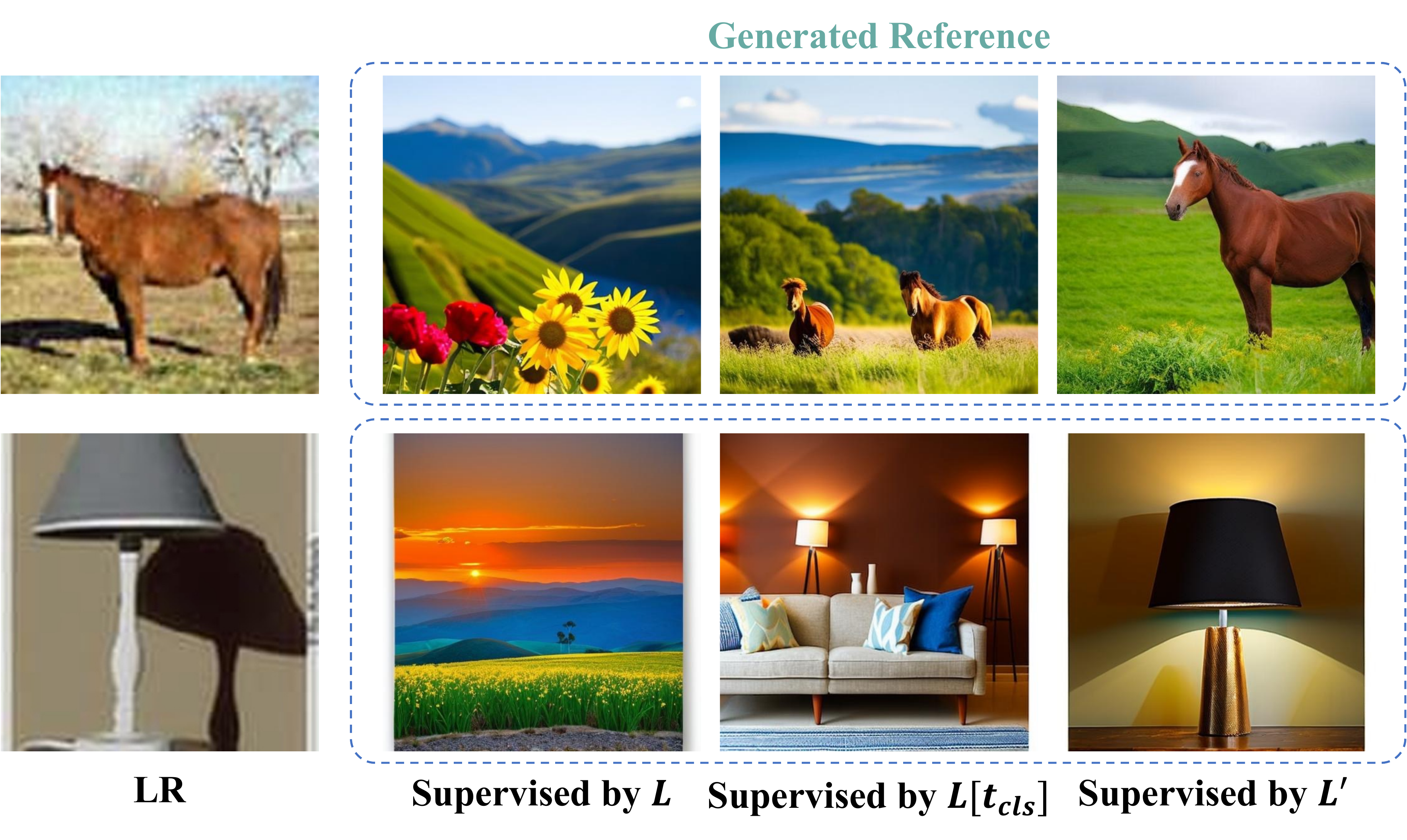}
    \vspace{-7mm}
    \caption{Reference images generated by cognitive encoders with different supervision methods.}\label{fig:supervision}
    % \vspace{-2mm}
\end{figure}

We observe that employing $\bm{L}$ directly as supervision for the cognitive embedding $\bm{E}$ (setting $T_e = T_l$) hinders the acquisition of cognitive information, as depicted in Figure~\ref{fig:supervision}. In this scenario, the generated reference images might prove irrelevant to low-resolution (LR) images. This limitation stems from the variability in caption length, which leads to Q-Former's learnable queries inadequately capturing semantic information at corresponding positions. To mitigate this issue, we propose using the last $T_e$ tokens for supervision for two reasons. Firstly, the last $T_e$ tokens in the CLIP text embedding inherently encapsulate an overarching representation of all preceding words~\cite{clip}, facilitated by the causal attention mechanism. This mitigates the requirement for a strict one-to-one correspondence between query ordering and semantic representation, thus enabling more effective learning by the queries. Secondly, within the supervision target $\bm{L^{\prime}}$, the last query consistently aligns with the class token, thereby preserving the full representational capacity of the class token. Compared to the direct utilization of $\bm{L}\left[t_{cls}\right]$ or single class token, our approach of employing the last $T_e$ tokens for supervision presents a more accurate understanding of LR images, which is supported by both Figure.~\textcolor{red}{5} in the main paper and Figure~\ref{fig:supervision}.

\begin{table}[!tp]
\begin{center}
\scalebox{0.9}{
\begin{tabular}{c|c}
\toprule
Number of Queries & Gen-score$\uparrow$ \\ \midrule
$T_e=30$          & 0.5983              \\
$T_e=40$          & 0.6048              \\
$T_e=50$          & \textbf{0.6147}     \\
$T_e=60$          & 0.6110              \\
$T_e=77$          & 0.6082              \\ 
\bottomrule
\end{tabular}
% \vspace{-1mm}
}
\caption{Reference image quality assessment using different numbers of learnable queries.}\label{tab:token_num}
\vspace{-7mm}
\end{center}
\end{table}

\begin{figure*}[t]
    \centering
    \includegraphics[width=0.95\textwidth]{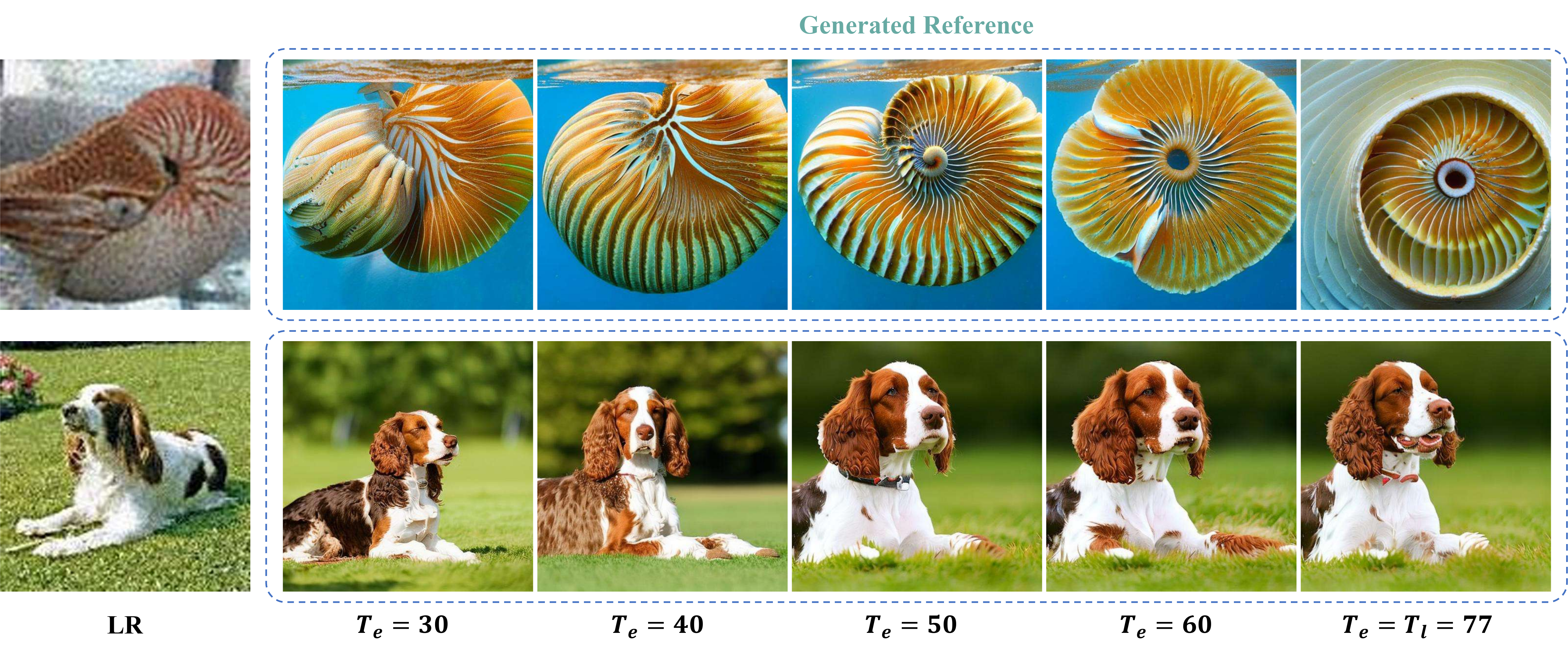}
    \vspace{-2mm}
    \caption{Reference images generated by cognitive encoders with different numbers of learnable queries.}\label{fig:cog_super_num}
    \vspace{-2mm}
\end{figure*}

% We investigated the inference of the number of learnable queries $T_e$ in our cognitive encoder through the generated reference images. We randomly selected 200 LRs from the ImageNet dataset and generated reference images for them using cognitive encoders with different numbers of learnable queries. We start with $T_e=30$ and increase $T_e$ in intervals of $10$ until the number of learnable tokens matches the number of CLIP language tokens $T_l=77$. Table~\ref{tab:token_num} shows that our cognitive encoder achieves the best generation performance when $T_e=50$. So we set $T_e=50$ in our cognitive encoder by default. We find that the quality of the generated images decreases instead after $T_e>50$ (as shown in Fig.~\ref{t_e}), probably because the excessive number of tokens increases the learning difficulty. $T_e=50$ strikes a good balance between cognitive accuracy and learning difficulty.

We investigate the impact of the number of learnable queries, denoted as $T_e$, in our cognitive encoder on the generation of high-quality reference images. This analysis involve the examination of 200 randomly selected low-resolution test images by varying the query number from 30 to 77. It is noted that the setting of $T_e=77$ in $\bm{L^{\prime}}$ differs from using $\bm{L}$ for supervision. This distinction arises from the fact that the final tokens of $\bm{L^{\prime}}$ are expanded with class tokens when the caption is not sufficiently lengthy. The results presented in Table~\ref{tab:token_num} demonstrate that our cognitive encoder achieves optimal performance when $T_e=50$ (where ``Gen-score'' is defined in the main paper). Hence, we establish $T_e=50$ as the default value. Notably, the quality of the generated images begins to decline for $T_e>50$, as also evidenced in Figure~\ref{fig:cog_super_num}. This decline might be attributed to increased learning complexity associated with a higher number of tokens.

\begin{figure}[t]
    \centering
    \includegraphics[width=0.33\textwidth]{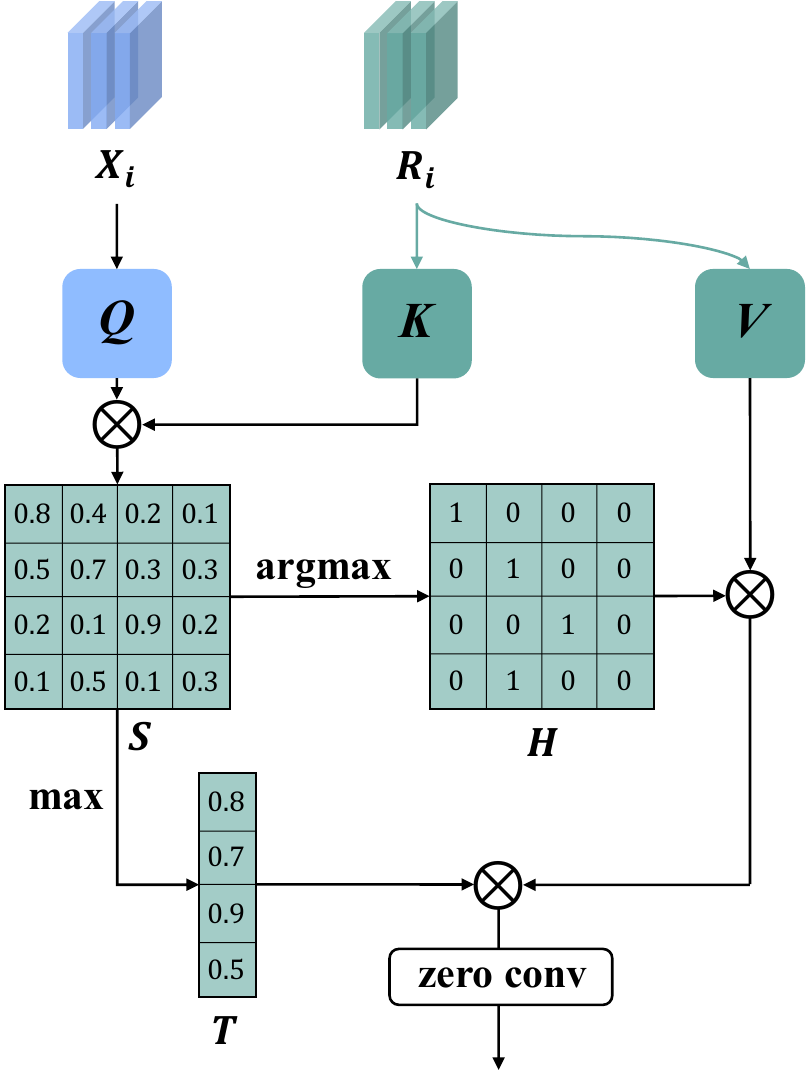}
    \caption{The architecture of one-hot reference attention in the All-in-Attention (AiA) module.}\label{fig:aia}
    \vspace{-6mm}
\end{figure}

\begin{figure*}[!htp]
    \centering
    \includegraphics[width=\textwidth]{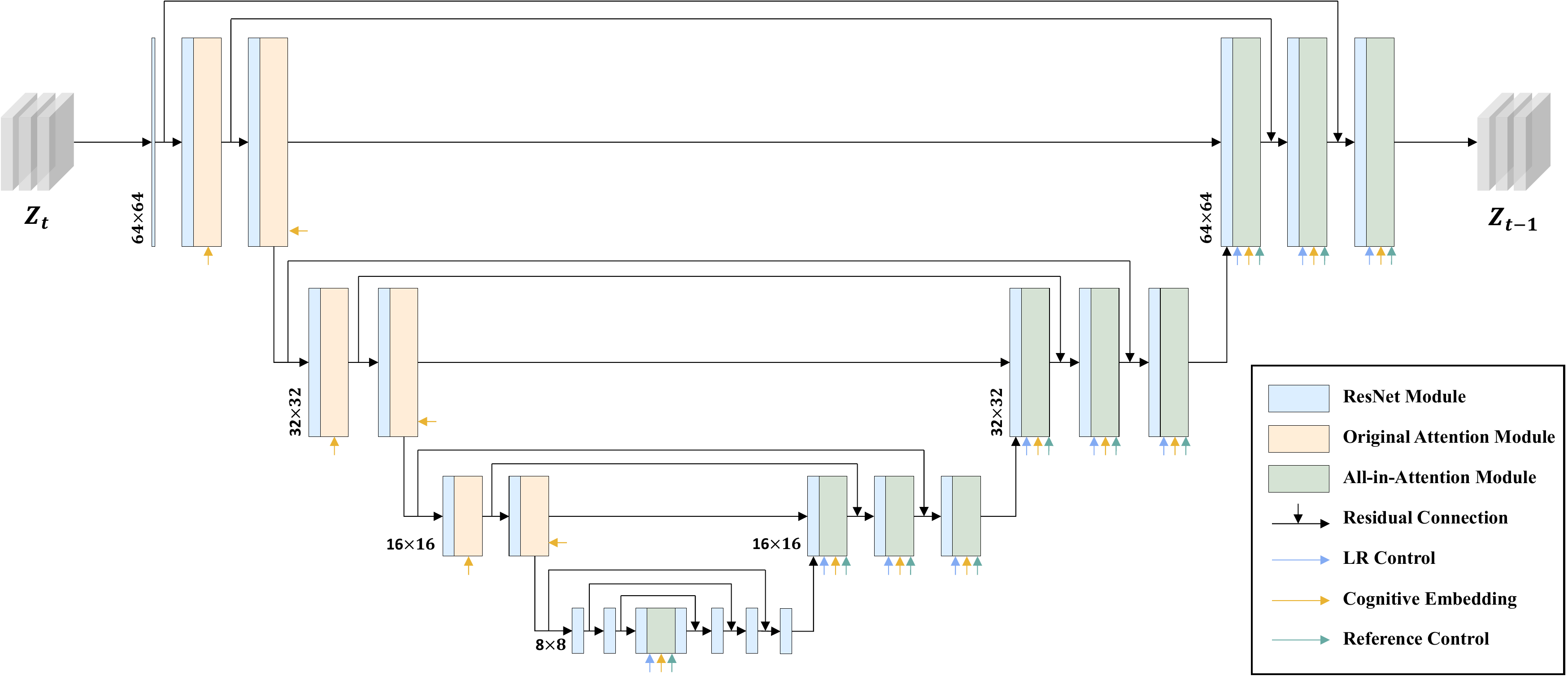}
    \caption{Network structure of the denoising U-Net in the proposed CoSeR framework.}\label{fig:detailed_unet}
    \vspace{-2mm}
\end{figure*}

\subsection{One-Hot Reference Attention} 
The reference image contains high-definition textures that maintain consistent semantics with the corresponding LR image. However, not all features from the reference image are useful for LR recovery. The conventional attention mechanism calculates the weighted sum of all queries in value features, potentially leading to a blurring effect~\cite{ttsr}. To address this issue, we introduce one-hot attention in the reference module to enhance the LR image with the most pertinent reference feature.

The one-hot attention mechanism is depicted in Figure~\ref{fig:aia}, where $\bm{Q}$, $\bm{K}$, and $\bm{V}$ denote the query, key, and value features, respectively. We represent the LR control and reference image control at the $i$-th scale as $\bm{X}_i$ and $\bm{R}_i$. $\bm{Q}\in \mathbb{R}^{B \times T_x \times C}$ and $\bm{K}, \bm{V}\in \mathbb{R}^{B \times T_r \times C}$ are derived from $\bm{X}_i, \bm{R}_i$. The similarity $\bm{S}\in \mathbb{R}^{B \times T_x \times T_r}$ between $\bm{Q}$ and $\bm{K}$ is computed with normalized inner product:
\begin{equation}\label{onehot_attention_1}
\begin{aligned}
\bm{S}=\left\langle \bm{Q}, \bm{K} \right\rangle.
\end{aligned}
\end{equation}
We derive the one-hot map $\bm{H}\in \mathbb{R}^{B \times T_x \times T_r}$ along the $T_r$ dimension of $\bm{S}$ and record the maximum values as $\bm{T}\in \mathbb{R}^{B \times T_x}$. The final output of the one-hot attention is then expressed as:
\begin{equation}\label{onehot_attention_1}
\begin{aligned}
\bm{Z}_{\text{out}} = {\rm ZeroConv} \left[ \left(\bm{H}\bm{V}\right) \odot \bm{T} \right],
\end{aligned}
\end{equation}
where $\odot$ denotes element-wise multiplication. It is noteworthy that we opt not to use \textit{softmax} and, instead, employ the correlation matrix $\bm{T}$ to diminish less similar features while amplifying those that are potentially valuable. Additionally, to prevent the newly introduced attention components from influencing the well-established representation of Stable Diffusion~\cite{stablediffusion} during early training, we integrate zero convolutions~\cite{controlnet} at the end. 

\begin{figure}[!htp]
    \centering
    \includegraphics[width=0.49\textwidth]{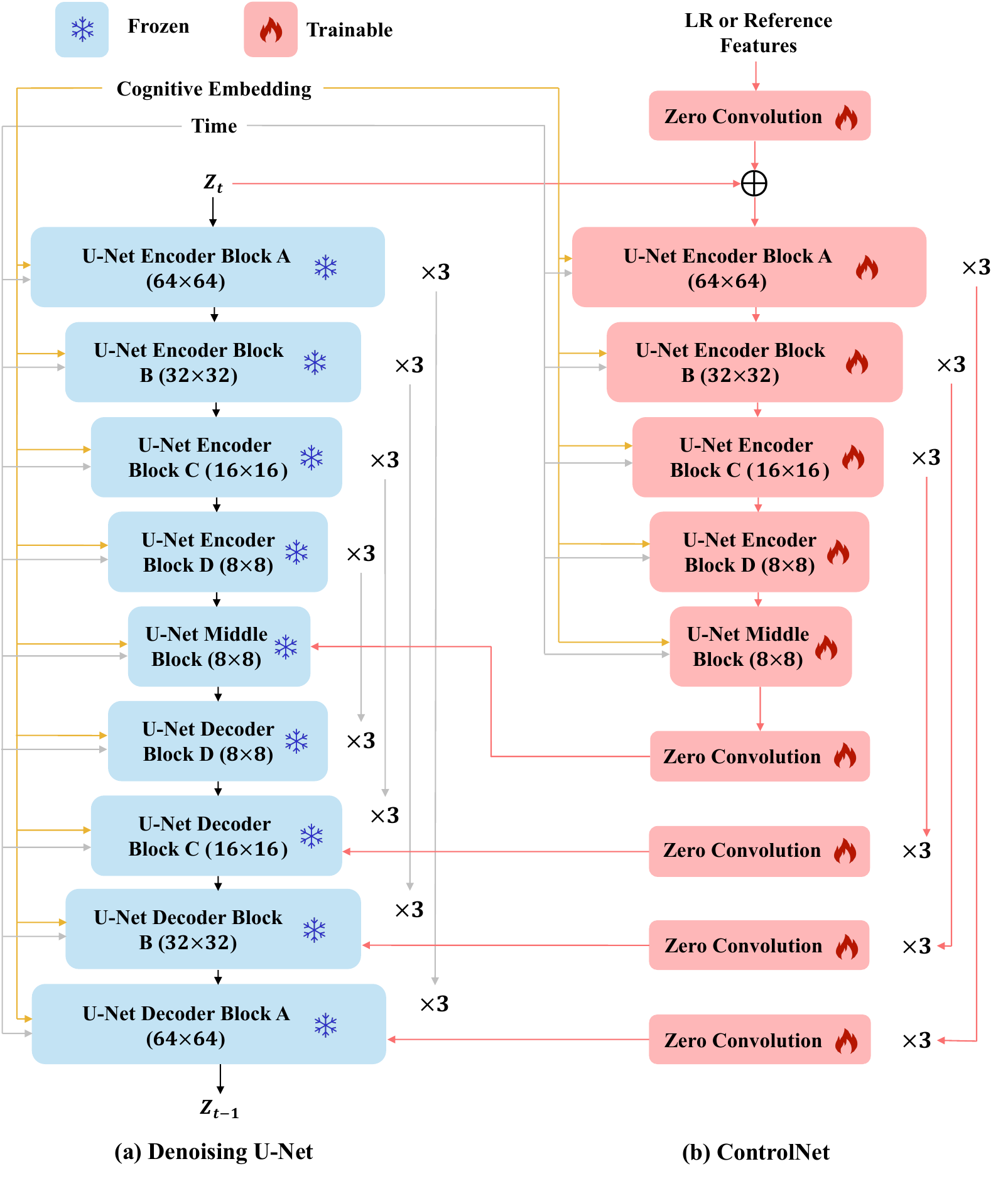}
    \caption{Network structure of ControlNet in the proposed CoSeR framework.}
    \label{fig:detailed_controlnet}
    \vspace{-4mm}
\end{figure}

\subsection{Network Structure} 

\noindent \textbf{Denoising U-Net.} The denoising U-Net in the proposed Cognitive Super-Resolution (CoSeR) network is depicted in Figure~\ref{fig:detailed_unet}. In our architecture, we adopt the All-in-Attention (AiA) module, replacing all original attention modules present in both the middle and decoder components of the Stable Diffusion denoising U-Net. It is crucial to highlight that cognitive embedding is utilized across all attention modules in the denoising U-Net, extending beyond solely the AiA modules.

\noindent \textbf{ControlNet.} We utilize ControlNet~\cite{controlnet} to generate multi-scale control features for both LR and reference images. As illustrated in Figure~\ref{fig:detailed_controlnet}, we mirror the weights and structure of the denoising U-Net in the ControlNet. Following~\cite{controlnet}, zero convolutions are incorporated at the beginning and end of the ControlNet module. Subsequently, the resulting control features are directed to the All-in-Attention (AiA) modules situated within the middle and decoder components of the denoising U-Net, excluding U-Net Decoder Block D, which lacks attention modules. Importantly, cognitive embedding is also employed in the ControlNet module.

\section{Additional Experiments}\label{sec2}

\subsection{Quantitative Comparisons to Official Models}

\begin{table*}[t]
\begin{center}

\scalebox{0.89}{
\begin{tabular}{c|c|ccccccc|c}
\toprule
Datasets                                                                      & Metrics              & RealSR & Real-ESRGAN+ & SwinIR-GAN & BSRGAN      & FeMaSR       & DiffBIR         & StableSR     & CoSeR           \\ \hline\hline
\multirow{5}{*}{\begin{tabular}[c]{@{}c@{}}ImageNet \\ Test2000\end{tabular}} & FID$\downarrow$      & 86.36  & 39.37        & 44.86      & 49.94       & 45.19        & $-$             & \underline{ 24.70}  & \textbf{19.41}  \\
                                                                              & DISTS$\downarrow$    & 0.2649 & 0.1915       & 0.2000     & 0.2043      & 0.1995       & $-$             & \underline{ 0.1608} & \textbf{0.1482} \\
                                                                              & LPIPS$\downarrow$    & 0.4519 & 0.3122       & 0.3327     & 0.3401      & 0.3403       & $-$             & \underline{ 0.2979} & \textbf{0.2863} \\
                                                                              & CLIP-Score$\uparrow$ & 0.6242 & 0.7642       & 0.7325     & 0.7126      & 0.7272       & $-$             & \underline{ 0.8459} & \textbf{0.8755} \\
                                                                              & MUSIQ$\uparrow$      & 50.18  & 61.92        & 57.60      & \underline{ 64.37} & 60.27        & $-$             & 63.20        & \textbf{65.51}  \\ \midrule
\multirow{5}{*}{RealSR~\cite{realsr}}                                                       & FID$\downarrow$      & 157.85 & 106.24       & 105.99     & 111.25      & 106.08       & \underline{ 90.30}     & 96.39        & \textbf{80.82}  \\
                                                                              & DISTS$\downarrow$    & 0.2529 & 0.2021       & 0.1969     & 0.2081      & 0.2125       & 0.1932          & \underline{ 0.1899} & \textbf{0.1826} \\
                                                                              & LPIPS$\downarrow$    & 0.3672 & 0.2805       & 0.2755     & 0.2801      & 0.2688       & 0.2967          & \underline{ 0.2639} & \textbf{0.2438} \\
                                                                              & CLIP-Score$\uparrow$ & 0.7458 & 0.8425       & 0.8425     & 0.8304      & 0.8473       & 0.8414          & \underline{ 0.8531} & \textbf{0.8545} \\
                                                                              & MUSIQ$\uparrow$      & 60.40  & 66.68        & 65.93      & 68.35       & 67.51        & 69.20           & \underline{ 69.25}  & \textbf{70.29}  \\ \midrule
\multirow{5}{*}{DRealSR~\cite{drealsr}}                                                      & FID$\downarrow$      & 148.58 & 97.60        & 98.94      & 110.53      & 95.71        & 86.49           & \underline{ 83.36}  & \textbf{71.22}  \\
                                                                              & DISTS$\downarrow$    & 0.2673 & 0.2121       & 0.2056     & 0.2033      & 0.2016       & \textbf{0.1959} & 0.2034       & \underline{ 0.1977}    \\
                                                                              & LPIPS$\downarrow$    & 0.4212 & 0.2973       & 0.2946     & 0.3062      & \underline{ 0.2777} & 0.3075          & 0.2960       & \textbf{0.2702} \\
                                                                              & CLIP-Score$\uparrow$ & 0.7360 & 0.8623       & 0.8571     & 0.8498      & 0.8680       & 0.8630          & \underline{ 0.8729} & \textbf{0.8766} \\
                                                                              & MUSIQ$\uparrow$      & 54.28  & 66.30        & 66.74      & 67.64       & 67.60        & 68.64           & \underline{ 69.57}  & \textbf{70.18}  \\ \bottomrule
\end{tabular}
}
\vspace{-1mm}
\caption{Quantitative comparisons to officially released models on both ImageNet Test2000 and real-world benchmarks RealSR and DRealSR. 
% The best results are highlighted in \textbf{bold}.
The best results are highlighted in \textbf{bold} and the second best results are in \underline{underlined}.
\vspace{-1mm}
}\label{tab:main_comparison_offi}
\end{center}
\end{table*}

\begin{table*}[t]
\begin{center}

\scalebox{0.89}{
\begin{tabular}{c|c|ccccccc}
\toprule
Datasets                                                                      & Methods & DISTS$\downarrow$ & LPIPS$\downarrow$ & CLIP-Score$\uparrow$ & PSNR$\uparrow$ & SSIM$\uparrow$  & FID$\downarrow$ & MUSIQ$\uparrow$ \\ \hline\hline
\multirow{2}{*}{\begin{tabular}[c]{@{}c@{}}ImageNet \\ Test2000\end{tabular}} & DiffBIR & 0.1523            & 0.3156            & 0.8683               & 21.12          & 0.5366          & 21.30           & \textbf{67.40}  \\
                                                                              & CoSeR   & \textbf{0.1482}   & \textbf{0.2863}   & \textbf{0.8755}      & \textbf{22.28} & \textbf{0.5998} & \textbf{19.41}  & 65.51           \\ \midrule
\multirow{2}{*}{RealSR}                                                       & DiffBIR & 0.1907            & 0.2727            & 0.8379               & 20.49          & 0.5511          & \textbf{78.31}  & 68.63           \\
                                                                              & CoSeR   & \textbf{0.1826}   & \textbf{0.2438}   & \textbf{0.8545}      & \textbf{21.24} & \textbf{0.6109} & 80.82           & \textbf{70.29}  \\ \midrule
\multirow{2}{*}{DRealSR}                                                      & DiffBIR & 0.2008            & 0.2980            & 0.8581               & 19.85          & 0.4934          & \textbf{68.21}  & 68.60           \\
                                                                              & CoSeR   & \textbf{0.1977}   & \textbf{0.2702}   & \textbf{0.8766}      & \textbf{19.95} & \textbf{0.5350} & 71.22           & \textbf{70.18}  \\ \bottomrule
\end{tabular}
}
\vspace{-1mm}
\caption{Quantitative comparisons between CoSeR and re-trained DiffBIR. 
% The best results are highlighted in \textbf{bold}.
The better results are highlighted in \textbf{bold}.
\vspace{-2mm}
}\label{tab:retrained_diffbir}
\end{center}
\end{table*}

For fair comparisons, we conduct a re-training of real-world super-resolution (SR) models using the ImageNet~\cite{imagenet} dataset in the main paper. Remarkably, our CoSeR model achieves the highest performance. To provide a comprehensive analysis, we compare CoSeR against the officially released models: RealSR~\cite{realsr_method}, Real-ESRGAN+~\cite{real-esrgan}, SwinIR-GAN~\cite{swinir}, BSRGAN~\cite{bsrgan}, FeMaSR~\cite{femasr}, DiffBIR~\cite{diffbir}, and StableSR~\cite{stablesr}. It is noted that we exclude the comparison with DiffBIR on the ImageNet Test2000 dataset due to potential data overlap with its official training set. Additionally, all diffusion-based models, including LDM, DiffBIR, StableSR, and CoSeR, employ 200 sampling steps. As outlined in Table~\ref{tab:main_comparison_offi}, across various evaluation metrics such as FID~\cite{fid}, DISTS~\cite{dists}, LPIPS~\cite{lpips}, CLIP-Score~\cite{clip}, and MUSIQ~\cite{musiq}, our method consistently excels, positioning CoSeR as the superior and more robust approach.

\subsection{Comparisons to Re-trained DiffBIR}
As an extension to the quantitative comparisons provided above, we further conduct a comparative analysis with DiffBIR~\cite{diffbir}, specifically re-trained using our ImageNet training set. The results displayed in Table~\ref{tab:retrained_diffbir} underscore the superiority of CoSeR across three benchmarks, demonstrating better performance across nearly all metrics.

\begin{figure}[!t]
    \centering
    \includegraphics[width=0.49\textwidth]{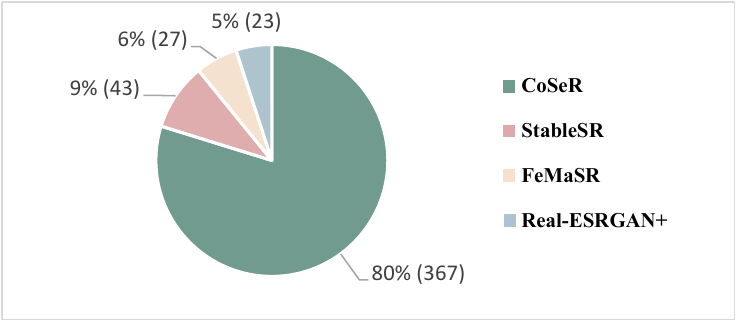}
    \caption{The voting results obtained from 23 users. The percentage of votes chosen along with the corresponding numerical count are adjacent to the pie chart.}
    \label{fig:user_study}
    \vspace{-2mm}
\end{figure}

\subsection{Voting Results of User Study}
As detailed in the main paper, we invite 23 subjects to discern the visually superior result among the four SR candidates generated by Real-ESRGAN+, FeMaSR, StableSR, and CoSeR. This user study encompasses 20 real-world low-resolution images sourced from the Internet or captured via mobile phones, resulting in a total of $20 \times 23$ votes gathered. The depicted voting results in Figure~\ref{fig:user_study} unequivocally illustrate the superior performance of our CoSeR.

\subsection{Number of Reference Images}

We investigate the influence of using multiple generated reference images on the quality of SR results. Employing the same LR input, we randomly sample noise maps to create several reference images utilizing identical cognitive embeddings. The findings presented in Table~\ref{tab:multiple_ref} demonstrate that introducing a greater number of reference images yields improved performance. However, it's noteworthy that the improvement plateaus when using 2 or 3 reference images, suggesting that these images already contain sufficient high-definition textures to guide the process. As a result, we recommend utilizing 2 reference images as an optimal balance between quality enhancement and computational efficiency.

\begin{table}[t]
\begin{center}

\scalebox{0.89}{
\begin{tabular}{c|ccc}
\toprule
Number of Ref. Images & FID$\downarrow$ & MUSIQ$\uparrow$ & MANIQA$\uparrow$ \\ \midrule
1                   & 19.80           & 64.21           & 0.2107           \\
2                   & \textbf{19.54}           & 64.85           & 0.2169           \\
3                   & 19.58           & \textbf{64.92}           & \textbf{0.2170}           \\
%4                   & 19.56           & 64.96           & 0.2170           \\ 
\bottomrule
\end{tabular}
}
\vspace{-1mm}
\caption{Results of using multiple reference images.
% The best results are highlighted in \textbf{bold}.
% The better results are highlighted in \textbf{bold}.
\vspace{-2mm}
}\label{tab:multiple_ref}
\end{center}

\end{table}

\begin{table}[t]
\begin{center}

\scalebox{0.8}{
\begin{tabular}{c|c|cccc}
\toprule
Datasets                                                                      & Metrics        & \begin{tabular}[c]{@{}c@{}}Real-\\ ESRGAN+\end{tabular} & FeMaSR & StableSR     & CoSeR       \\ \hline\hline
\multirow{2}{*}{\begin{tabular}[c]{@{}c@{}}ImageNet \\ Test2000\end{tabular}} & PSNR$\uparrow$ & \textbf{22.64}                                          & 20.95  & 22.24        & \underline{ 22.28} \\
                                                                              & SSIM$\uparrow$ & \textbf{0.6268}                                         & 0.5674 & \underline{ 0.6093} & 0.5998      \\ \midrule
\multirow{2}{*}{RealSR}                                                       & PSNR$\uparrow$ & \textbf{21.93}                                          & 20.45  & 21.19        & \underline{ 21.24} \\
                                                                              & SSIM$\uparrow$ & \textbf{0.6497}                                         & 0.6061 & \underline{ 0.6247} & 0.6109      \\ \midrule
\multirow{2}{*}{DRealSR}                                                      & PSNR$\uparrow$ & \textbf{20.47}                                          & 18.46  & 19.86        & \underline{ 19.95} \\
                                                                              & SSIM$\uparrow$ & \textbf{0.5781}                                         & 0.5036 & \underline{ 0.5487} & 0.5350      \\ \bottomrule
\end{tabular}
}
\vspace{-1mm}
\caption{Pixel-level PSNR and SSIM assessment of SR quality.
% The best results are highlighted in \textbf{bold}.
% The better results are highlighted in \textbf{bold}.
\vspace{-4mm}
}\label{tab:pixel_iqa}
\end{center}

\end{table}

\subsection{Pixel-level Image Quality Assessment}
% As shown in Table~\ref{tab:pixel_iqa}, we evaluate the recovery quality of Real-ESRGAN+, FeMaSR, StableSR and CoSeR based on pixel-level IQA metrics such as PSNR and SSIM. Although, our method is significantly less effective than methods that are more biased towards pixel-level alignment, like Real-ESRGAN+. However, thanks to the AiA module, both the PSNR in Table~\ref{tab:pixel_iqa} and the comparison of PSNR and SSIM with DiffBIR in Table~\ref{tab:retrained_diffbir} demonstrate that our method has comparable or even better pixel-level fidelity than other diffusion-based methods.

While acknowledging that PSNR and SSIM metrics exhibit a weak correlation with human perception, particularly for large-scale super-resolution tasks, we present the corresponding results in Table~\ref{tab:pixel_iqa} for reference purposes. Our CoSeR achieves favorable results. The All-in-Attention (AiA) module contributes to competitive or superior pixel-level fidelity compared to other diffusion-based models like DiffBIR and StableSR, as shown in Table~\ref{tab:retrained_diffbir} and Table~\ref{tab:pixel_iqa}.

\section{Qualitative Comparisons}\label{sec3}
\begin{figure*}[!t]
    \centering
    \includegraphics[width=0.93\textwidth]{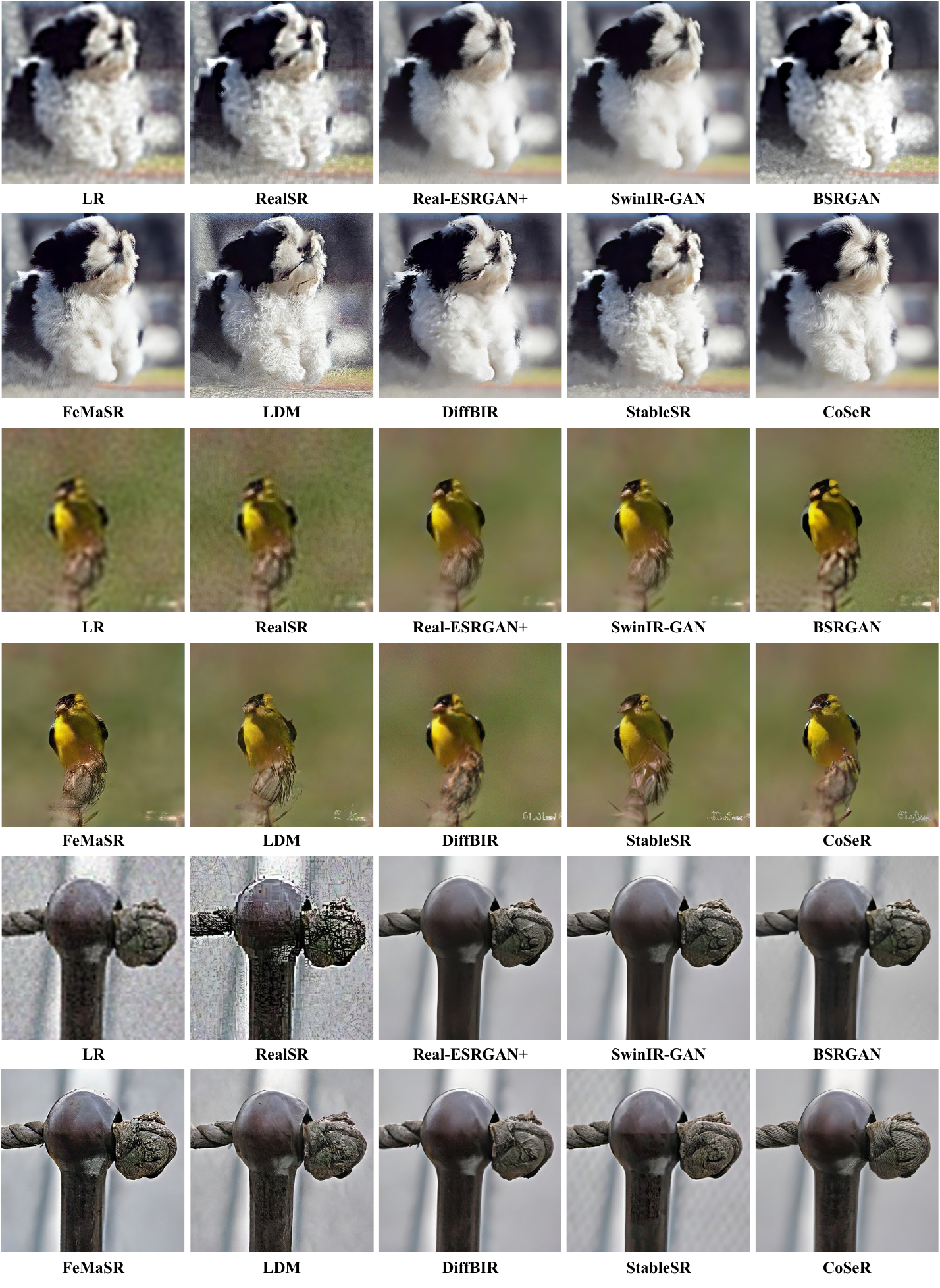}
    \vspace{-2mm}
    \caption{Qualitative comparisons on ImageNet Test2000 dataset (part 1/4).}
    \label{fig:supp_imagenet1}
    % \vspace{-2mm}
\end{figure*}

\begin{figure*}[!t]
    \centering
    \includegraphics[width=0.93\textwidth]{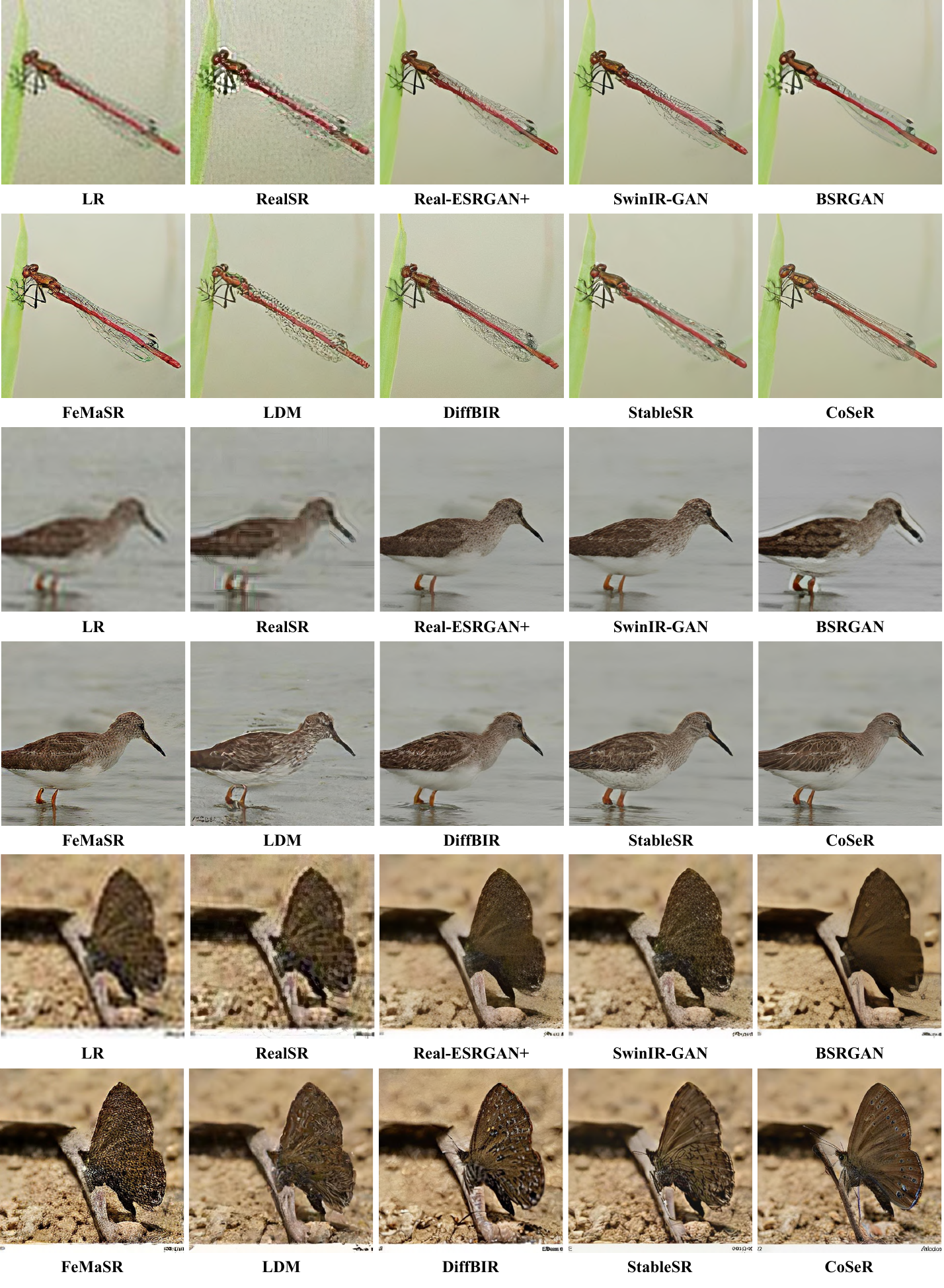}
    \vspace{-2mm}
    \caption{Qualitative comparisons on ImageNet Test2000 dataset (part 2/4).}
    \label{fig:supp_imagenet2}
    % \vspace{-2mm}
\end{figure*}

\begin{figure*}[!t]
    \centering
    \includegraphics[width=0.93\textwidth]{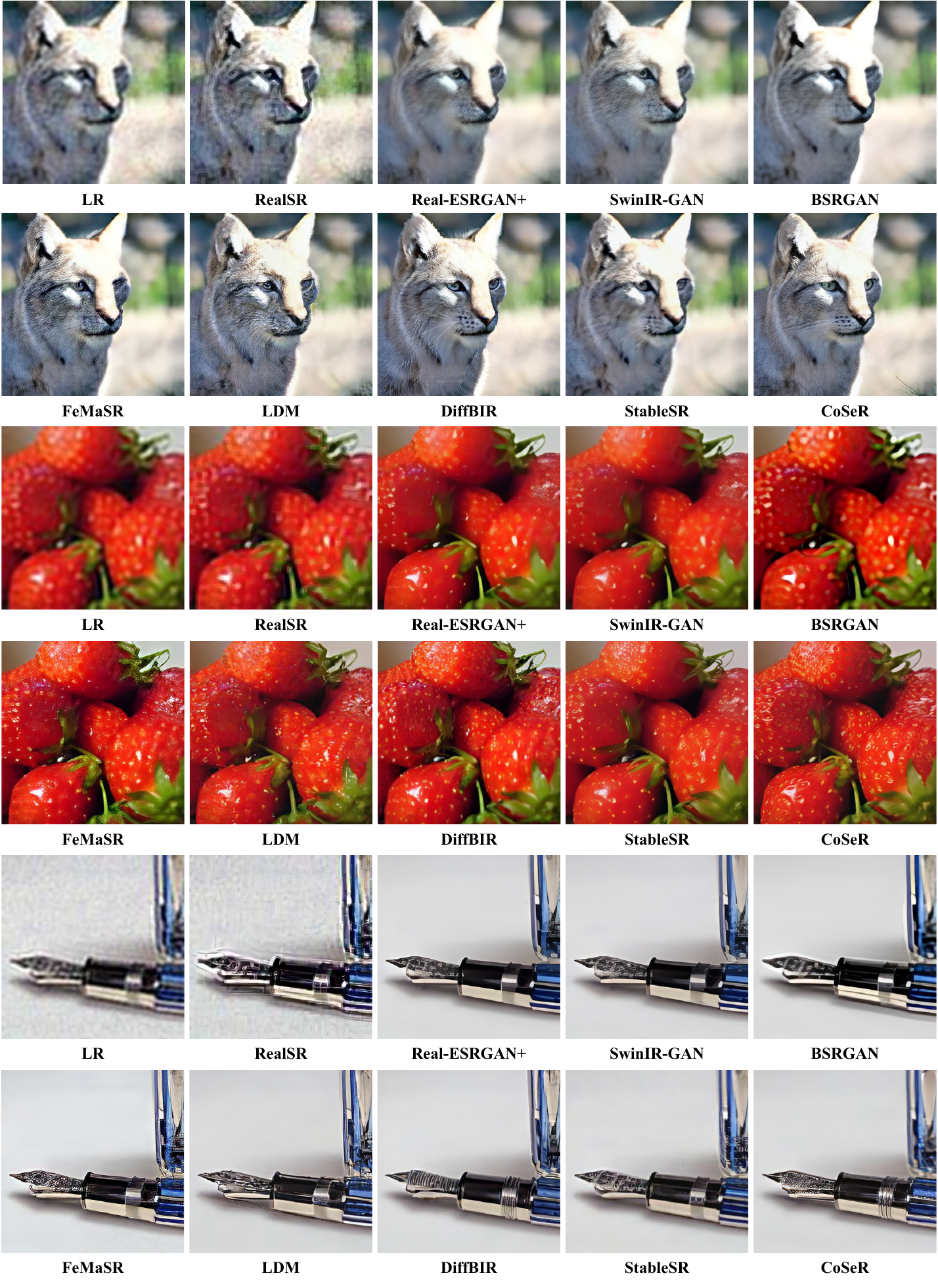}
    \vspace{-2mm}
    \caption{Qualitative comparisons on ImageNet Test2000 dataset (part 3/4).}
    \label{fig:supp_imagenet3}
    % \vspace{-2mm}
\end{figure*}

\begin{figure*}[!t]
    \centering
    \includegraphics[width=0.93\textwidth]{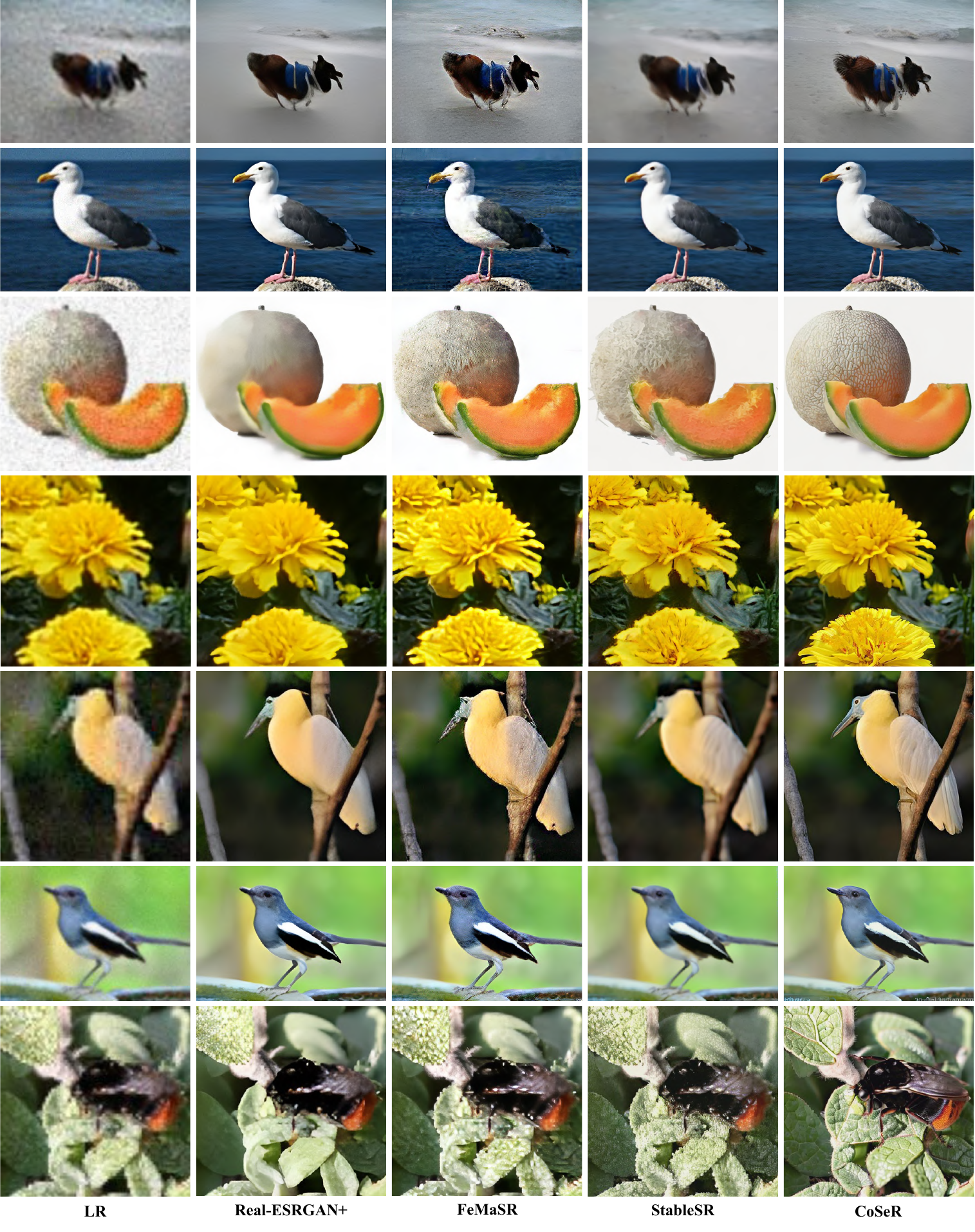}
    \vspace{-2mm}
    \caption{Qualitative comparisons on real-world or unknown degradation type images (part 1/2).}
    \label{fig:supp_realworld1}
    % \vspace{-2mm}
\end{figure*}

\begin{figure*}[!t]
    \centering
    \includegraphics[width=0.93\textwidth]{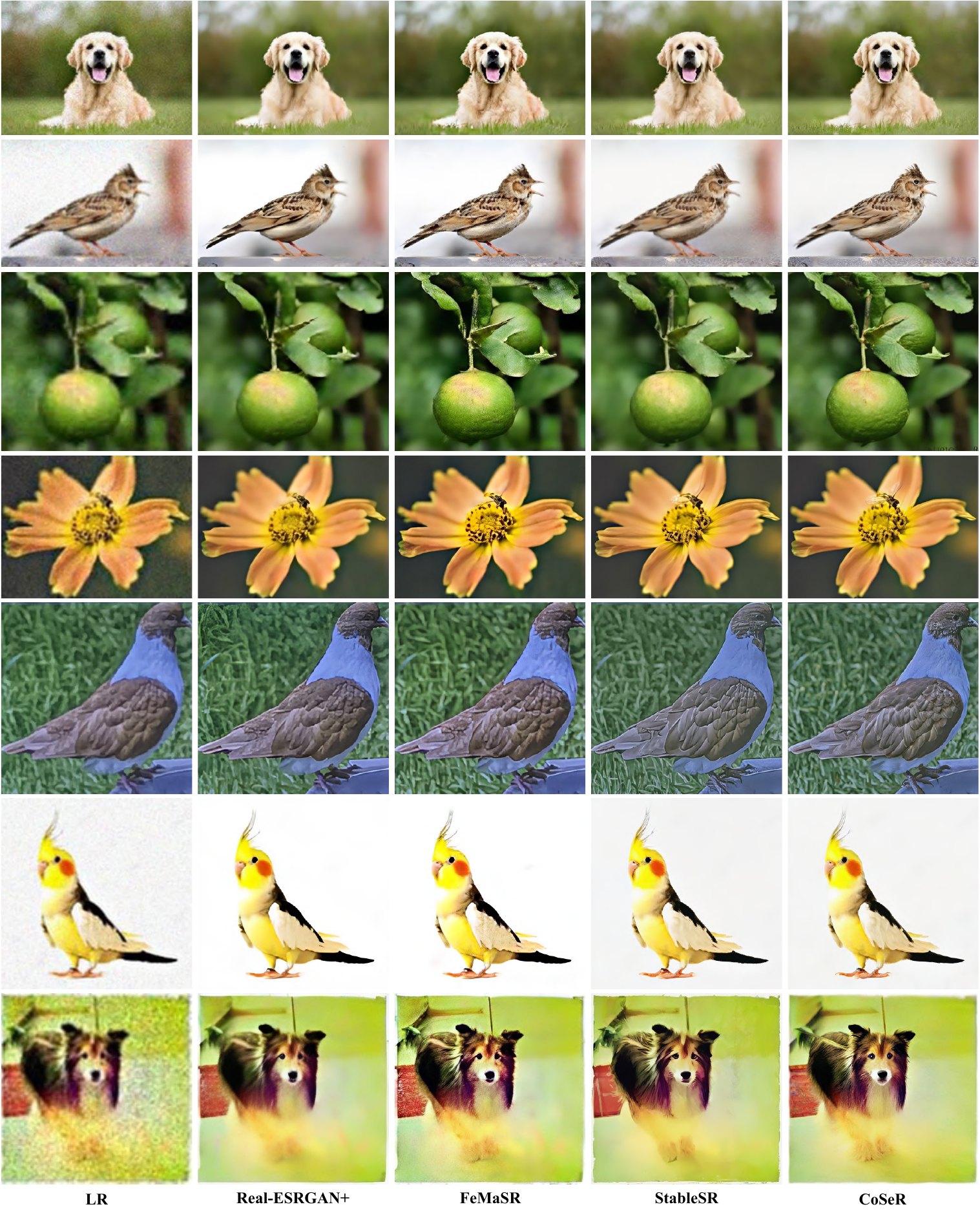}
    \vspace{-2mm}
    \caption{Qualitative comparisons on real-world or unknown degradation type images (part 2/2).}
    \label{fig:supp_realworld2}
    % \vspace{-2mm}
\end{figure*}

\begin{figure*}[!t]
    \centering
    \includegraphics[width=0.93\textwidth]{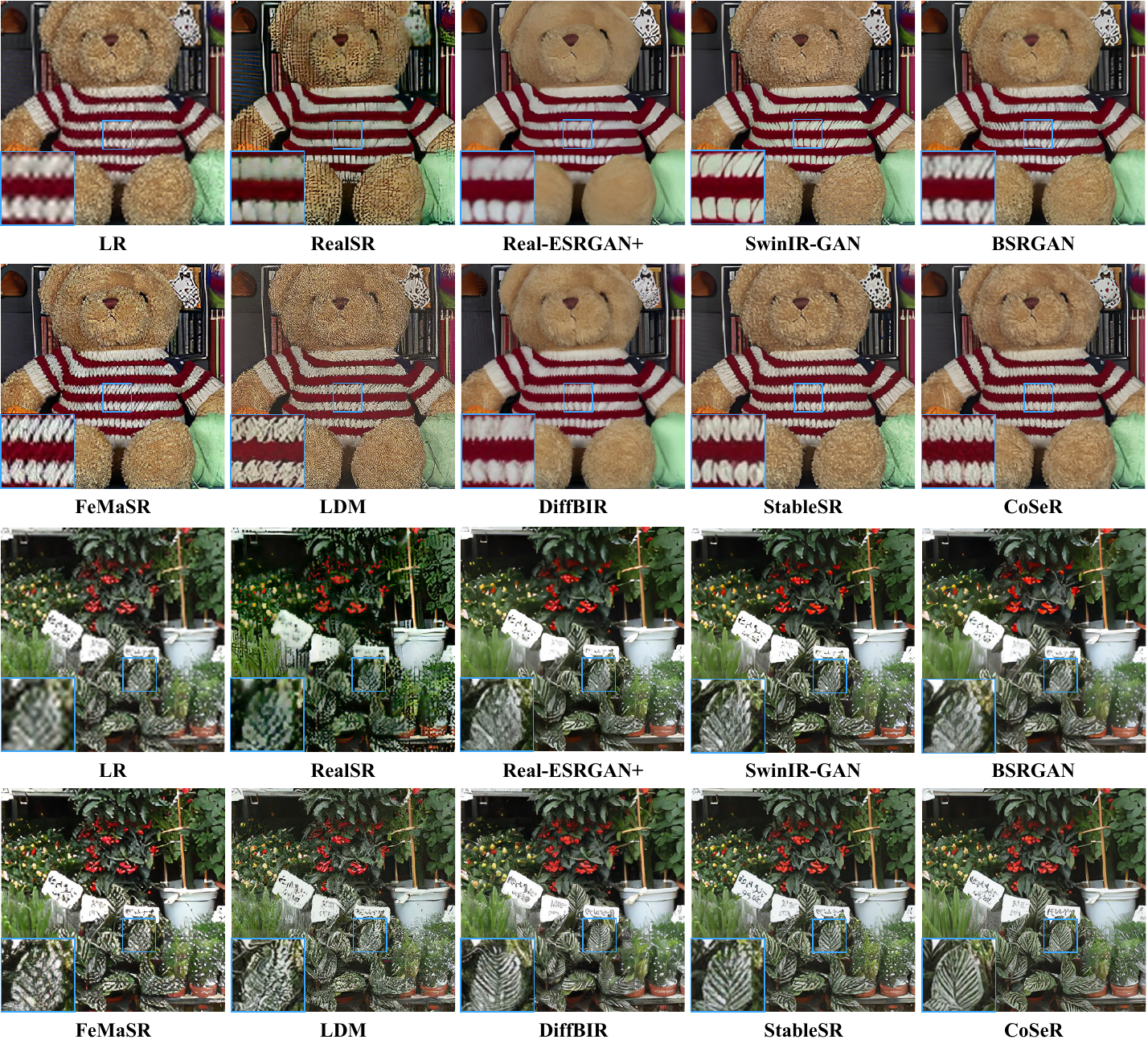}
    \vspace{-2mm}
    \caption{Qualitative comparisons on RealSR and DRealSR datasets.}
    \label{fig:supp_drealsr_realsr}
    % \vspace{-2mm}
\end{figure*}

We provide visual comparisons on ImageNet Test2000 dataset (Figures~\ref{fig:supp_imagenet1}, \ref{fig:supp_imagenet2}, \ref{fig:supp_imagenet3}), real-world or unknown degradation type images (Figures~\ref{fig:supp_realworld1}, \ref{fig:supp_realworld2}), RealSR and DRealSR datasets (Figure~\ref{fig:supp_drealsr_realsr}). Our CoSeR obtains outstanding visual performance.

\section{Future Work}\label{sec4}
The cognitive-based recovery process extends beyond super-resolution (SR) tasks; it is beneficial for various visual tasks such as deblurring, denoising, and inpainting. Our future work includes expanding its application to more diverse image restoration tasks. Additionally, prevalent SR algorithms based on diffusion models often require a large number of sampling steps for higher visual quality. Addressing the challenge of accelerating the sampling process without compromising SR performance is also a focal point of our ongoing research.

% \begin{figure*}[!t]
%     \centering
%     \includegraphics[width=0.93\textwidth]{supp_imagenet4.pdf}
%     \vspace{-2mm}
%     \caption{Qualitative comparisons on ImageNet Test2000 dataset (part 4/4).}
%     \label{fig:supp_imagenet4}
%     % \vspace{-2mm}
% \end{figure*}

% \begin{figure*}[!t]
%     \centering
%     \includegraphics[width=0.93\textwidth]{supp_drealsr.pdf}
%     \vspace{-2mm}
%     \caption{Qualitative comparisons on DRealSR dataset.}
%     \label{fig:supp_drealsr}
%     % \vspace{-2mm}
% \end{figure*}

\clearpage
\clearpage

{
    \small
    \bibliographystyle{ieeenat_fullname}
    \bibliography{main}
}

% \input{X_suppl}

\end{document}